\documentclass[conference]{IEEEtran}
\usepackage[utf8]{inputenc}
\usepackage{url}
\usepackage{color}
\usepackage{amsmath,amsthm,amssymb,amscd} 
\usepackage{graphicx} 
  \usepackage{psfrag,  caption}
    \usepackage{subfig}
\usepackage{verbatim}
\usepackage{enumerate}
\DeclareMathOperator{\ima}{im}

\IEEEoverridecommandlockouts
\begin{document}
\author{
    \IEEEauthorblockN{Yu-Min Chung\IEEEauthorrefmark{1},  Chuan-Shen Hu\IEEEauthorrefmark{2}, Austin Lawson\IEEEauthorrefmark{1}, Clifford Smyth\IEEEauthorrefmark{1}\thanks{Clifford Smyth was supported by the Simons Foundation grant 360468.}}\\
 \IEEEauthorblockA{\IEEEauthorrefmark{1}\textit{Department of Mathematics and Statistics}\\
		\textit{University of North Carolina at Greensboro}\\
		Greensboro, North Carolina 27412, USA \\
		\{y\_chung2, azlawson, cdsmyth\}@uncg.edu}\\
\IEEEauthorblockA{\IEEEauthorrefmark{2}\textit{Department of Mathematics}\\
		\textit{National Taiwan Normal University}\\
		Taipei City 106, Taiwan \\
		peterbill26@hotmail.com}

}

\title{TopoResNet: A hybrid deep learning architecture and its application to skin lesion classification}

\IEEEpeerreviewmaketitle
\maketitle


\begin{abstract}
Skin cancer is one of the most common cancers in the United States. As technological advancements are made, algorithmic diagnosis of skin lesions is becoming more important.  In this paper, we develop algorithms for segmenting the actual diseased area of skin in a given image of a skin lesion, and for classifying different types of skin lesions pictured in a given image.  The cores of the algorithms used were based in persistent homology, an algebraic topology technique that is part of the rising field of Topological Data Analysis (TDA).  The segmentation algorithm utilizes a similar concept to persistent homology that captures the robustness of segmented regions.  For classification, we design two families of topological features from persistence diagrams---which we refer to as {\em persistence statistics} (PS) and {\em persistence curves} (PC), and use linear support vector machine as classifiers. { We also combined those topological features, PS and PC, into ResNet-101 model, which we call {\em TopoResNet-101}, the results show that PS and PC are effective in two folds---improving classification performances and stabilizing the training process.   Although convolutional features are the most important learning targets in CNN models, global information of images may be lost in the training process. Because topological features were extracted globally, our results show that the global property of topological features provide additional information to machine learning models.
}
\end{abstract}

\begin{figure}
\subfloat[MEL]{\includegraphics[scale=0.1]{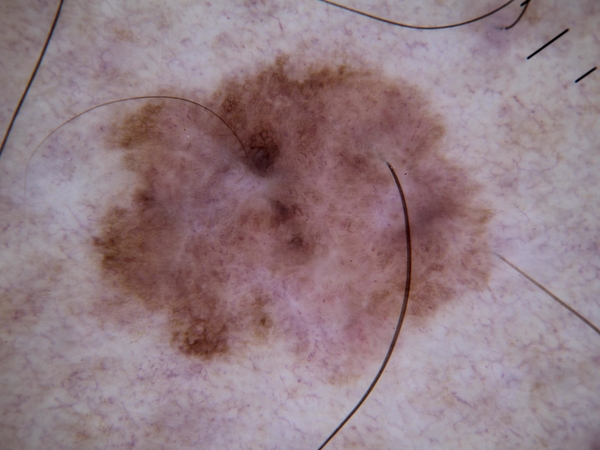}} \hfill
\subfloat[NV]{\includegraphics[scale=0.1]{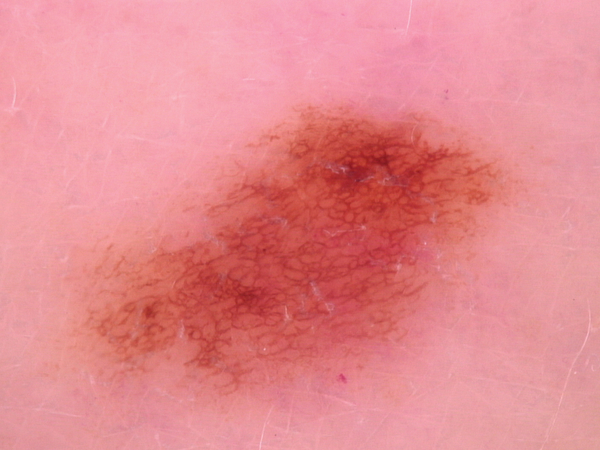}} \hfill
\subfloat[BCC]{\includegraphics[scale=0.1]{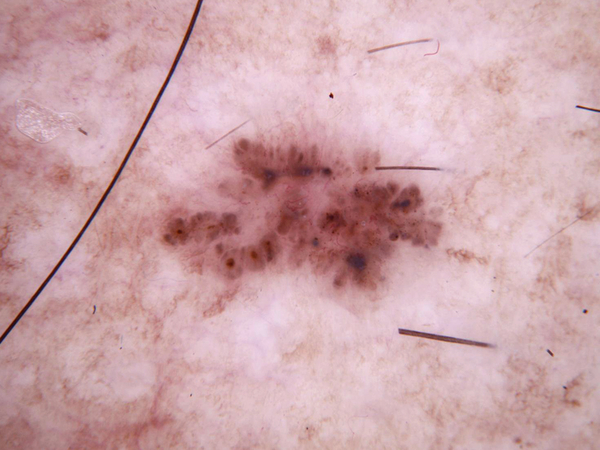}} \hfill
\subfloat[AKIEC]{\includegraphics[scale=0.1]{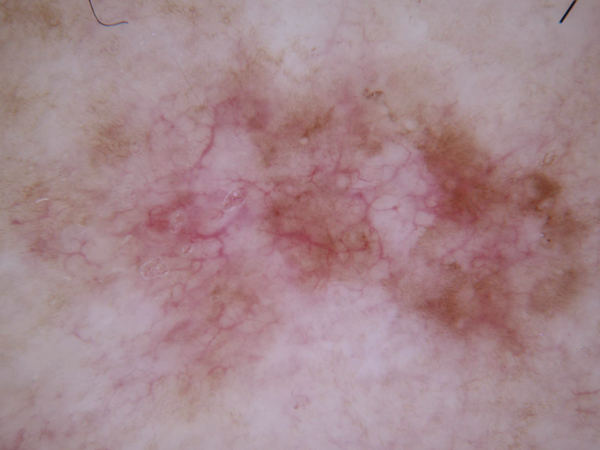}}\\
\subfloat[BKL]{\includegraphics[scale=0.1]{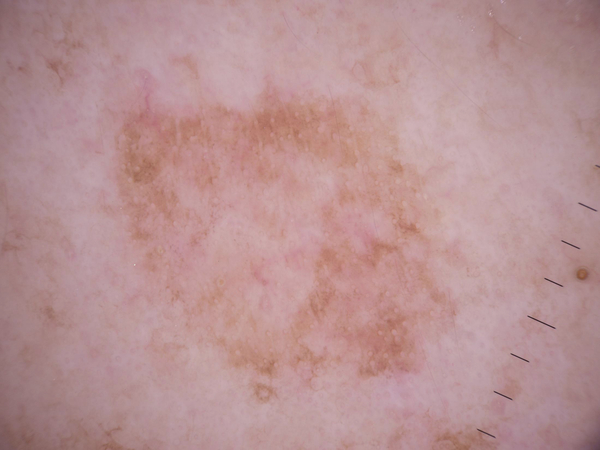}} \hfill
\subfloat[DF]{\includegraphics[scale=0.1]{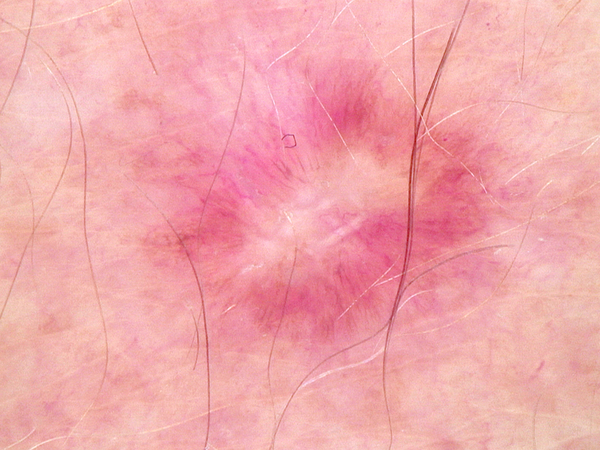}} \hfill
\subfloat[VASC]{\includegraphics[scale=0.1]{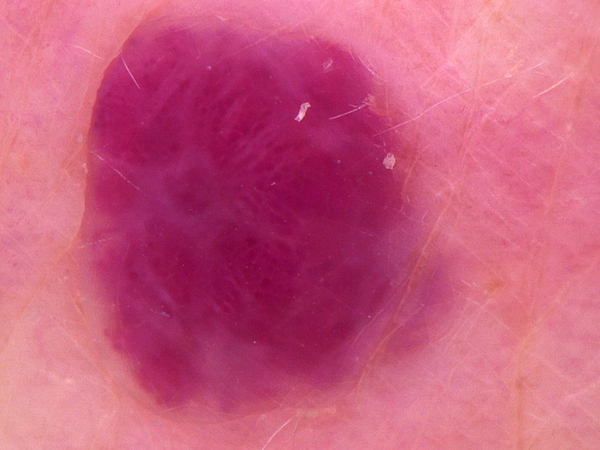}} \hfill
\caption{Sample images of different types of skin lesions from the ISIC training dataset. MEL is Melanoma, NV is Melanocytic nevus, BCC is Basal cell carcinoma,  AKIEC is Actinic keratosis, BKL is Benign keratosis,  DF is Dermatofibroma, and VASC is Vascular lesion.}
\label{fig:sample images}
\end{figure}

\section{Introduction}
The International Skin Imaging Collaboration (ISIC, \cite{ISIC}) has put forth a number of imaging challenges to the scientific community \cite{ISICchallenges,Tschandl2018,ISICdata}.  These challenges have presented unique opportunities for researchers to test novel computer vision ideas to improve the detection of skin cancer with the long-term goal of facilitating early treatment and greatly improving patient outcomes. This is a worthy goal. In the United States, the five-year survival rate for treated melanoma in the United States is 98\% among those with localized disease and 17\% among those in whom spread has occurred \cite{melanoma-survival}.

The ISIC 2018 challenge \cite{ISIC2018} was to design skin lesion diagnostic algorithms (and test their efficacy) using ISIC's archive of over 13000 dermatoscopic images collected from a variety of sources \cite{HAM10000}.
Each image was one of the following seven types: MEL, NV, BCC, AKIEC, BKL, DF, and VASC (see Fig.~\ref{fig:sample images} for sample images).
The algorithms were trained on a set containing 10015 images, and validated on a set of 193 images. The holdout set consisted of 1512 images and we report our scores from this set.  The algorithms we developed were based on the powerful tool of persistent homology \cite{edelsbrunner2000topological} and a novel concept that called persistence curves (PC) \cite{ChungLawson}, and persistence statistics (PS) \cite{ChungDayRBC}.  

Dermatologist-level diagnosis of skin lesions has been obtained via machine learning analysis of raw images in \cite{Esteva} (see \cite{Esteva} also for references to earlier approaches).  

{
The main contribution of this work has two folds.  First, we craft topological features based on PC and PS for the classification task.  Second, we design a novel deep learning architecture, called {\em TopoResNet-101}.  The essential idea is to combine the above topological features with those produced by ResNet-101.  The features generated by ResNet-101 are often local and geometric (e.g. gradients, edges) information. On the other hand, topological ones are global information, and hence, can be used as additional information for the original neural network model.  As shown in Section \ref{Sec: Top ResNet}, {TopoResNet-101} shows evidence on several advantages, such as accuracy and stability.  To the best of our knowledge, this work is the first one to   combine such topological features and those from ResNet-101 in the classification task.

The outline of this paper is as follows. In Section \ref{Sec : Topological Features}, we discuss those mathematical backgrounds in order to properly define PC and PS.  In Section \ref{Sec: Top ResNet}, we introduce the {TopoResNet-101} and topological rate $\alpha$. The main classification results are shown in Section~\ref{Sec:Exp} and the conclusion is in Section~\ref{sec:conclusion}. 

}

\section{Topological Features}
\label{Sec : Topological Features}
{The goal of this section is to extract PS and PC from skin images. Because these features are based on {\em persistence diagrams} and {\em persistent homology}, we review those mathematical backgrounds in \ref{sec:PH}.  The PC and PS as features will be presented in \ref{sec:PS and PC}.  As shown in Figure~\ref{fig:sample images}, typical images consist of lesions and non-lesions. In addition to feature engineering, we also propose an intuitive method in \ref{sec:seg} for segmenting the lesion part of the image.  There are deep learning methods to perform image segmentation, such as ~\cite{Fully-Net,U-Net}.  It would be interesting to explore those segmentation methods in the content of skin images, but it will be beyond the scope of this paper. The main focus of this paper is to design topological features and combine them with ResNet-101. We propose a segmentation algorithm in Section \ref{sec:seg} that is purely data-driven and only dependent on the topology and geometry of these skin lesion images. 
}

\begin{figure}
\centering
\includegraphics[width=\linewidth]{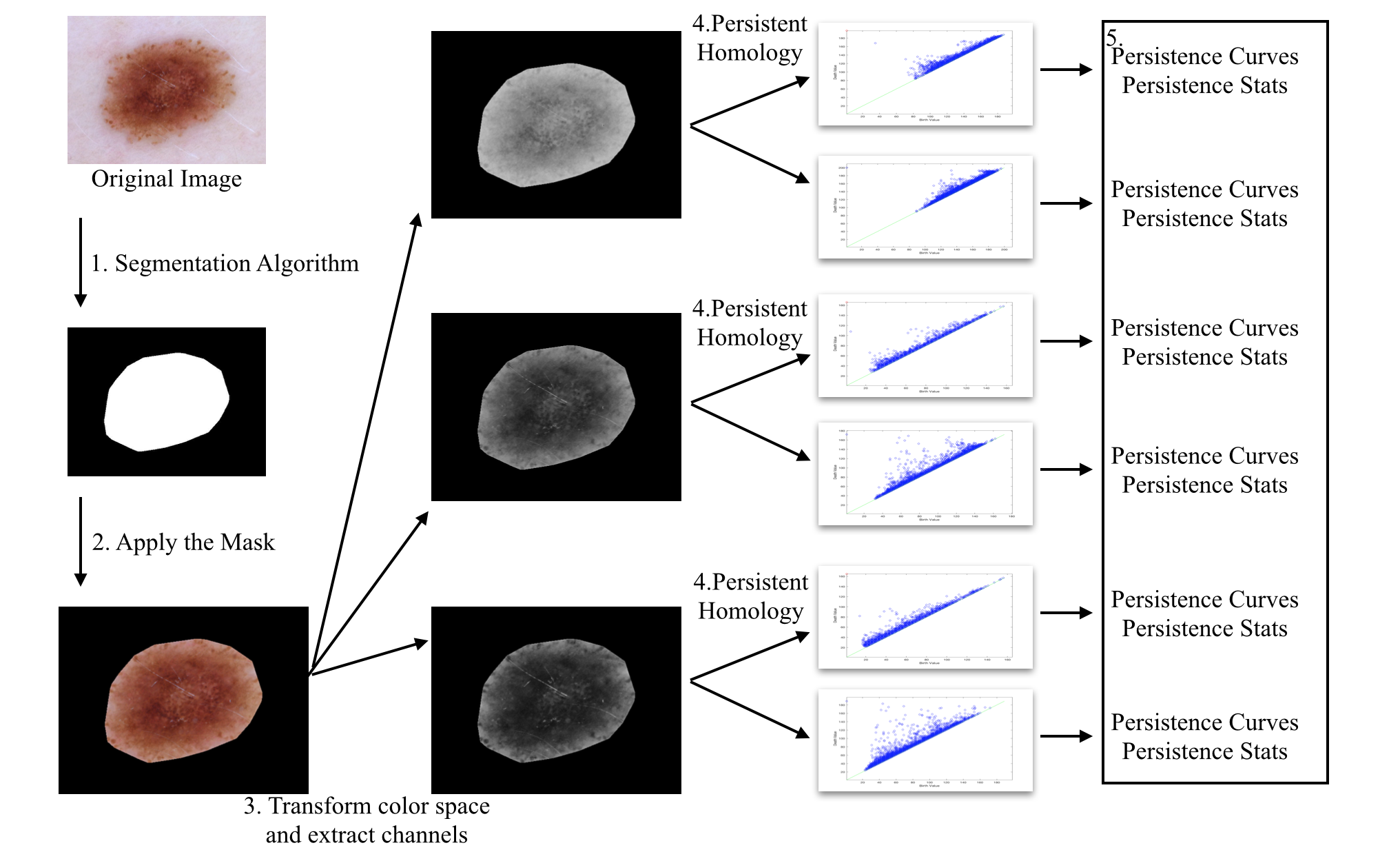}
\caption{Schematic pipeline of our proposed features extractions.  First, for each image, we apply the mask produced by the segmentation algorithm developed in Section~\ref{sec:seg}; second, for each mask image, we calculate its cubical persistent homology which we review in Section~\ref{sec:PH}; finally, from the persistence diagrams, we extract PS and PC, which are discussed in Section~\ref{sec:PS and PC}. }
\label{fig:general pipeline}
\end{figure}

\subsection{Persistent Homology}
\label{sec:PH}
Algebraic topology is a classical subject and has a long history within mathematics. {\it Persistent homology}, formally introduced in \cite{edelsbrunner2000topological}, brings the power of algebraic topology to bear on real world data. The field has proven useful in many applications, such as neuroscience \cite{bendich2016persistent}, medical biology \cite{li2015identification}, sensor networks \cite{de2007coverage}, social networks \cite{carstens2013persistent}, physics \cite{donato2016persistent}, computation \cite{lloyd2016quantum}, nanotechnology \cite{nakamura2015persistent}, natural language processes \cite{Zhu2013} and more. We'll give a brief overview of homology and persistent homology for images and refer the reader to \cite{kaczynski2004computational} and \cite{dlotko2016topological} for a more detailed exposition.

\begin{figure}
\centering
\includegraphics[scale=0.3]{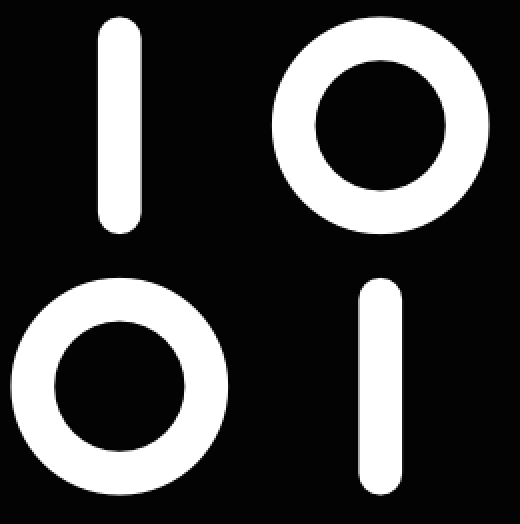}
\caption{The Betti numbers of an binary image. By convention a binary image represents the cubical complex $X$ of white pixels in the image, and its Betti number is $\beta_0(X) = 4$ and $\beta_1(X)=2$. 
Note that if the image is surrounded by a boundary of white pixels, then $\beta_0(X)=5$ and $\beta_1(X)=3$.}
\label{fig:betti numbers binary image}
\end{figure}

Informally, homology counts topological features such as connected components (0-dimensional homological features), holes (1-dimensional homological features), voids (2-dimensional homological features), and so on.  The counts of such $k$-dimensional holes are the well-known {\it Betti numbers}. In binary images, a black pixel is indicated by a value of $0$ and a white pixel by a value of $1$.  We interpret $0$- and $1$-dimensional homological features in binary images as follows.  We count connected clusters of white pixels as $0$-dimensional homological features and connected clusters of black pixels (surrounded by white pixels) as $1$-dimensional homological features (see Fig.~\ref{fig:betti numbers binary image} for an example).  Let $X$ be a binary image.  We denote the $k$th Betti numbers of $X$ by $\beta_k(X)$. To formalize this, we will treat binary images as cubical complexes, which we describe below.  We refer readers to \cite{kaczynski2004computational} for more detailed discussions on cubical complexes and homology.

 We consider intervals of the form $[\ell,\ell+1]$ or $[\ell,\ell] := [\ell] = \{\ell\}$ where $\ell\in\mathbb{Z}$, these are called \textit{elementary intervals}. Intervals of the form $[\ell]$ are called \textit{degenerate}. We define an \textit{elementary cube} to be a finite product of such intervals. In other words, $Q$ is an elementary cube if $Q = I_1\times I_2\times\ldots\times I_n$ where $I_j$ is an elementary interval for $j = 1,\ldots n$. The \textit{dimension} of $Q$, denoted $\mathrm{dim}(Q)$, is the number of non-degenerate intervals in the product. We say the set $X$ is \textit{cubical} if it can be written as a finite union of elementary cubes. Let $\mathcal{K}(X) = \{Q\in\mathcal{K}\mid Q\subset X\}$ denote the set of cubes making up $X$ and let $\mathcal{K}_k(X) = \{Q\in\mathcal{K}(X)\mid \mathrm{dim} Q = k\}$ denote the set of $k$-dimensional cubes in $X$. 

Now that we have a topological framework, we seek to provide a complementary algebraic framework. For this, we fix a ring $R$ and cubical set $X$. (Note: a ring \cite{DummitFoote} is a system of numbers which admits addition and multiplication e.g. $\mathbb{R}$, $\mathbb{Z}$ or $\mathbb{Z}_2 = \{ 0, 1 \}$.) We define the \textit{$k$-th chain module} over $X$, denoted $C_k(X)$ to be the formal span of its elementary cubes of dimension $k$. That is, \[C_k(X; R) = \left\{\sum_{Q\in\mathcal{K}_k(X)}\alpha_QQ \colon\alpha_Q\in R\right\}.\] 
For each $k\in \mathbb{N}$, $C_k(X;R)$ is an algebraic structure known as an $R$-module. If $R = \mathbb{R}$ (or more generally is a field) then the $R$-module $C_k(X;R)$ is just a vector space over $\mathbb{R}$.  It is worth mentioning that computing persistent homology using $R= \mathbb{Z}_2$ is much simpler and efficient than using $R=\mathbb{R}$, and, although the homologies are different in general, much of the utility is the same \cite{Zhu2013}.

Each element of a chain module is called a \textit{chain}. We now define the \textit{algebraic boundary map}, denoted $\partial$, between chain modules. First, for any interval $[\ell,\ell + 1]$, $\partial([\ell,\ell+1]) := [\ell+1] - [\ell]$ and $\partial([\ell]) = 0$ for every degenerate interval. Next, we define $\partial$ for elementary cubes. \[\partial(I_1\times I_2\times\ldots\times I_m) = \sum_{j=1}^m I_1\times \ldots \times \partial(I_j)\times\ldots I_m. \] Note that if $I_j = [\ell,\ell+1]$ we define the term \begin{align*}
I_1\times \ldots \times \partial(I_j)\times\ldots I_m &= I_1\times \ldots \times [\ell+1]\times\ldots I_m\\& - I_1\times \ldots \times [\ell]\times\ldots I_m
\end{align*} and if $I_j$ is degenerate, we define the term to be $0$. Finally, given a chain $c = \sum_{i=1}^m \alpha_iQ_i$ we set \[\partial(c) = \sum_{i=1}^m\alpha_i\partial(Q_i)\]
It is a well-known and important fact that $\partial\partial c = 0$. Note also that $\partial_k := \partial\mid_{C_k(X; R)}: C_k(X; R)\to C_{k-1}(X;R)$ is a map from the $k$-chain module to the $(k-1)$-chain module. Thus given a cubical set $X$ we can construct the \textit{chain complex} of $X$, denoted $\mathcal{C}(X;R)$, the collection of all $k$-chain modules together with their boundary maps. To avoid notation overload, we will write $C_k$ for $C_k(X;R)$ when the context is clear. We can visualize a chain complex through the following sequence of mappings
\[\ldots \stackrel{\partial_3}{\longrightarrow}C_2\stackrel{\partial_2}{\longrightarrow}C_1\stackrel{\partial_1}{\longrightarrow}C_0\stackrel{\partial_0}{\longrightarrow}0\]

The kernel, $\ker \partial_k := \{c \in C_k \ | \ \partial_k(c) = 0 \}$, of the $k$-th boundary map is called the set of \textit{$k$-cycles} or just \textit{cycles} and the image $\ima \partial_k := \{ \partial_{k}(c) | c \in C_{k} \}$ is the {\it $k+1$-boundary} or just the \textit{boundary}. Because $\partial\partial \equiv 0$ we $\ima\partial_k$ is a normal additive subgroup of $\ker\partial_{k+1}$. Thus we define the \textit{$k$-th homology group} of $\mathcal{K}$ to be the quotient group \cite{DummitFoote}
\[H_k(X;R) = \ker\partial_k/\ima\partial_{k+1}.\]

The \textit{$k$-th  Betti number} is defined to be the \textit{rank} (or \textit{order}) of the $k$-th homology group (its dimension as an $R$-module).  We denote this by $\beta_k(X;R)$.

In computer vision, a natural generation of binary images is grayscale image. Consider a grayscale image $I$ where each pixel value $I(x,y)$ is between $0$  and $255$. As $I(x,y)$ increases from $0$ to $255$ the pixel $(x,y)$ transitions from a color of black at $0$, to steadily lightening shades of gray, to pure white at $255$.

To obtain a binary image from $I$, one might threshold $I$ by some value $t$ to obtain a binary image $T(I,t)$.  The pixel function of $T(I,t)$ is $T(x,y,t)$ where $T(x,y,t) = 1$ if $I(x,y)\le t$ and $0$ otherwise. We alternately view $T(I,t)$ as the set of pixels $(x,y)$ for which $T(x,y,t)=1$. Thus $T(I,s) \subset T(I,t)$ if $s \leq t$. 

A user's choice of any given threshold $t$ at which to calculate the homologies of $T(I,t)$ would be arbitrary. Persistent homology offers a methodology to consider all possible threshold values on $I$ at once.

Suppose $X$ represents a cubical set. We define a \textit{filtration} of $X$ to be a sequence of cubical sets indexed by a finite discrete set (often of real numbers) \[X_{a_1}\subset X_{a_2}\ldots\subset X_{a_n} = X\]
where $a_i<a_j$ if $i<j$. It is also handy to sometimes set $X_0 = \emptyset$.  We can define a \textit{filtering function} $f:\mathcal{K}(X)\to \{a_1,\ldots, a_n\}$ that assigns each cube $C\in\mathcal{K}(X)$ to the complex in which it appears first. That is if $f(C) = a_k$ then $C\in \mathcal{K}(X_{a_k})$ and $C\notin \mathcal{K}(X_{a_{k-1}})$. Because of the inclusions, this also guarantees $C \in\mathcal{K}(X_{a_j})$ if and only if $j \geq f(C)$ . In particular, if $B$ is any cube in the boundary of $C$ then $f(B)\le f(C)$. Hence we can see $\mathcal{K}(X_{a_i}) = \{C\mid f(C)\le a_i\}$. This last definition is quite useful in the case of images. Recall that every binary image can be viewed as a cubical complex. Hence given a grayscale image $I$, we can create a filtering function for this image to send each elementary 2-cube and its boundary elements to the pixel value of that cube. Thus, an $n\times m$ grayscale image acts as a filtering function on the cubical set $[0,n-1]\times[0,m-1]$.

\begin{figure}
\centering
 \includegraphics[width=\linewidth]{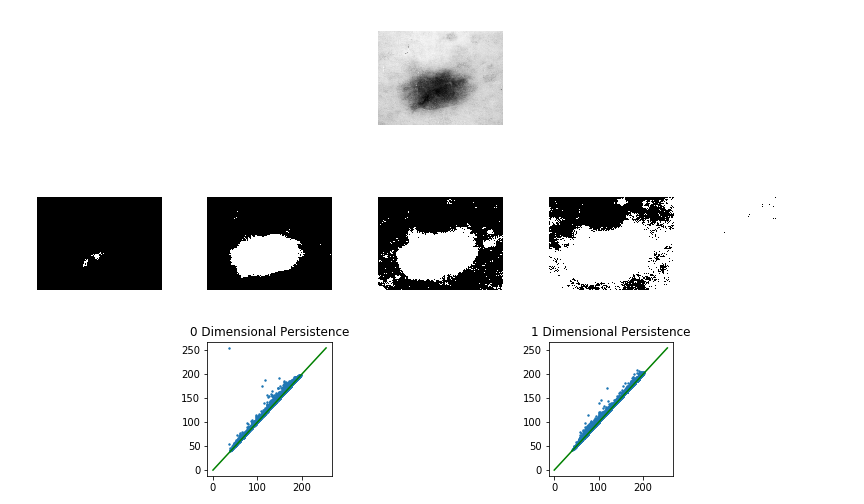}
 \caption{Top: An image from the ISIC dataset. Middle: A small sample of the full filtration of the top image. Bottom: The 0 and 1 dimensional persistence diagrams corresponding to the full filtration of the top image}\label{Filtration}
\end{figure}

Suppose we have a filtration of cubical sets, \[X_1\subset X_2\subset\ldots\subset X_n.\] The inclusion induces a homomorphism (see \cite{DummitFoote}) between the homology groups so that for each $k$ we have, \[H_k(X_1)\stackrel{f_1}{\longrightarrow} H_k(X_2)\stackrel{f_2}{\longrightarrow}\ldots \stackrel{f_{n-1}}{\longrightarrow} H_k(X_n)\] where each $f_i$ is a homomorphism. We say a homology class $\alpha$ is \textit{born} at $i$ if we have $\alpha\in H_k(X_i)$ and $\alpha\notin f_{i-1}(H_k(X_{i-1})$. We say $\alpha$ \textit{dies} at $H_k(X_i)$ if $\alpha\in H_k(X_{i-1})$ and one of the following hold: 
\begin{itemize}
\item $f_{i-1}(\alpha)$ is trivial; or
\item if $\alpha$ is born at $j$ and another class $\beta$ is born at $\ell < j$ and $f_{i-1}(\alpha) = f_{i-1}(\beta)$.
\end{itemize}
The last condition is described as the \textit{elder rule}, which allows us to uniquely define the death of a class. This rule says in the choice between two classes, we choose to keep the ``oldest'' class. We can guarantee that every homology class $\alpha\in H_k(X_i)$ for some $i$ has a birth time. We cannot guarantee that each class has a death time. For such classes, we assign the ``death time'' as $\infty$. This procedure allows us to define a unique multi-set of points, one point $(b,d)$ for each homology class where $b$ is the birth time of the class and $d$ is its death time. By collecting these pairs accounting for their multiplicity, we obtain a summary of the data called a persistence diagram for each dimension $k$. Refer to Fig.~\ref{Filtration} for an example of a filtration and corresponding persistence diagrams for an image.

It has been proven that the persistence diagram is a stable summary of a space in the sense that small changes in the space correspond to small changes in the corresponding diagram \cite{cohen2007stability} . These diagrams are an integral part of our algorithm as described in Section~\ref{Sec:Exp}

For a grayscale image $I$ we use the filtration $\{T(I,t)\}_{t=0}^{255}$. and calculate persistence diagrams for dimensions 0 and 1 (2 dimensional homology for 2-D images is trivial). We extend this notion to color images, by considering each channel in the color space individually. Suppose image $I$ has $I(x,y) = [R(x,y),G(x,y),B(x,y)]$ where $0 \leq R(x,y), G(x,y), B(x,y) \leq 255$ are the intensities of red, green, and blue light at pixel $(x,y)$, with $0$ being no intensity and $255$ being full intensity. Then one could obtain the persistence diagrams of $R(x,y)$, $B(x,y)$ and $G(x,y)$ viewed each as a separate grayscale image.

\subsection{Persistence Statistics and Persistence Curves}
\label{sec:PS and PC}
{Both PC and PS can be viewed as summaries of persistence diagrams. The main idea is to extract topological information from persistence diagrams, and use them as features to build the classification model. 
PCs were introduced in \cite{ChungLawson}, and are proven successful to texture datasets;  persistence statistics were studied in \cite{ChungDayRBC} to describe different types of human red blood cells.}

PS are statistical measurements of the birth and death coordinates.  Given a persistence diagram $P$, denote the birth and death coordinates by $b$ and $d$, respectively.  Recall that from Section \ref{sec:PH}, persistence diagrams are multi-sets of pairs of points, $(b,d)$, where $b\; (d)$ indicates the birth (death) value of a generator, respectively.  Since 2D images are considered, there are two persistence diagrams , $P_0$ and $P_1$, associated with each image.  $P_0$ contains information about 0-dimensional holes, i.e. connected components; $P_1$ contains information about 1-dimensional holes.   In this work, we consider only those nontrivial pairs, i.e. $(b,d)$ where $b < d$, and hence, $P$ contains finite number of pairs of points.  The widely-used quantity, $d-b$, represents the ``life'' of this generator, meaning how persist this generator is in the filtration.   One of the simplest summaries of the persistence diagram is called {\it total persistence}, and is defined as $L_i = \sum_{(b,d)\in P_i} d-b$, where $i=0,~1$. We also consider the midlife coordinates, $M_i = \{ (b+d)/2,\;(b,d)\in P_i \}$, and normalized lifespan $p_i = \{ (d-b)/L_i, \; (b,d)\in P_i \}$, where $i=0, \;1$.  $M_i$ reveals information about the locations of generators in persistence diagrams while $p_0$ reveals information about robustness of generators.  Notice that $M_i$ and $p_i$ may be viewed as empirical distributions.  Our first set of features for the classification is the standard statistics measurements to describe those distributions.   In particular, the PS we used in this work were
\begin{enumerate}
\item means of $M_i$, and $p_i$;
\item standard deviations of $M_i$, and $p_i$;
\item skewness of $M_i$, and $p_i$;
\item kurtosis of $M_i$, and $p_i$;
\item medians of $M_i$, and $p_i$;
\item $25$-th and $75$-th percentiles of $M_i$, and $p_i$;
\item interquartile ranges of $M_i$, and $p_i$;
\end{enumerate}
where $i=0,1$.  As shown in Table \ref{tab:sample PS}, they are samples of persistence statistics used in the article.
In addition to above statistics measurements, we also consider so called {\it persistence entropy} introduced in \cite{PersistentEntropy}, which is defined as $-\sum_{(b,d)\in P_i} p\log(p)$, for $i=0,1$.  Persistence entropy can be viewed as the diversity of the lifespans. 
\begin{table}\renewcommand{\arraystretch}{1.5}
\caption{Sample $M_0$ persistence statistics from the X channel of the XYZ color space of images in Fig.~\ref{fig:sample images}.}
\label{tab:sample PS}
\begin{tabular}{|c|c|c|c|c|c|c|}
\hline
Disease & mean & std & skewness & kurtosis & median & iqr  \\ \hline
MEL &    2.2533  &  1.6644 &   3.6107  &  3.0519 &   2.4897 &   2.0668   \\ \hline
NV &  2.6123 &   2.3389 &   2.0343  &  2.2425 &   2.7211 &   3.4245  \\ \hline
BCC &   6.8147  &  3.0709  &  3.2841 &   2.1271 &  10.9705 &   2.7159  \\ \hline
AKIEC & 3.3722 &   3.3452  &  3.8697 &   2.7465  &  4.4496 &   6.8388  \\ \hline
BKL & 4.2876 &   2.7614  &  3.7254  &  7.5341 &   3.5813 &   4.6003    \\ \hline
DF & 1.8916  &  6.4557  &  1.9783 &   3.4724  &  2.3310  &  2.8247   \\ \hline
VASC &    2.5901 &   2.0824  &  4.6341  &  2.7230 &   2.5502 &   1.8615 \\ \hline
\end{tabular}
\end{table}

\begin{figure}
\subfloat[$\beta_0(t)$]{\includegraphics[scale=0.1]{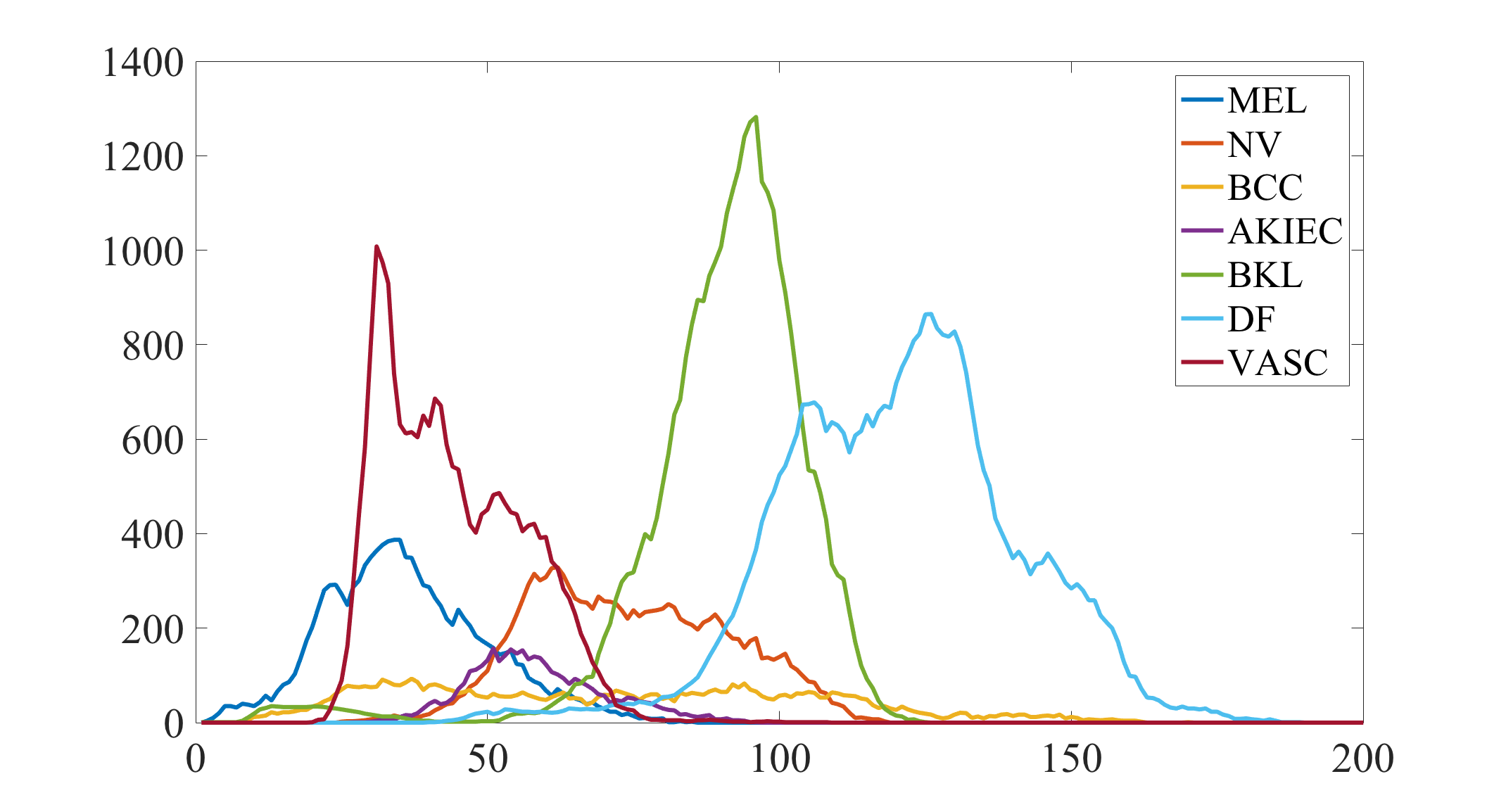}}\\
\subfloat[$E_0(t)$]{\includegraphics[scale=0.1]{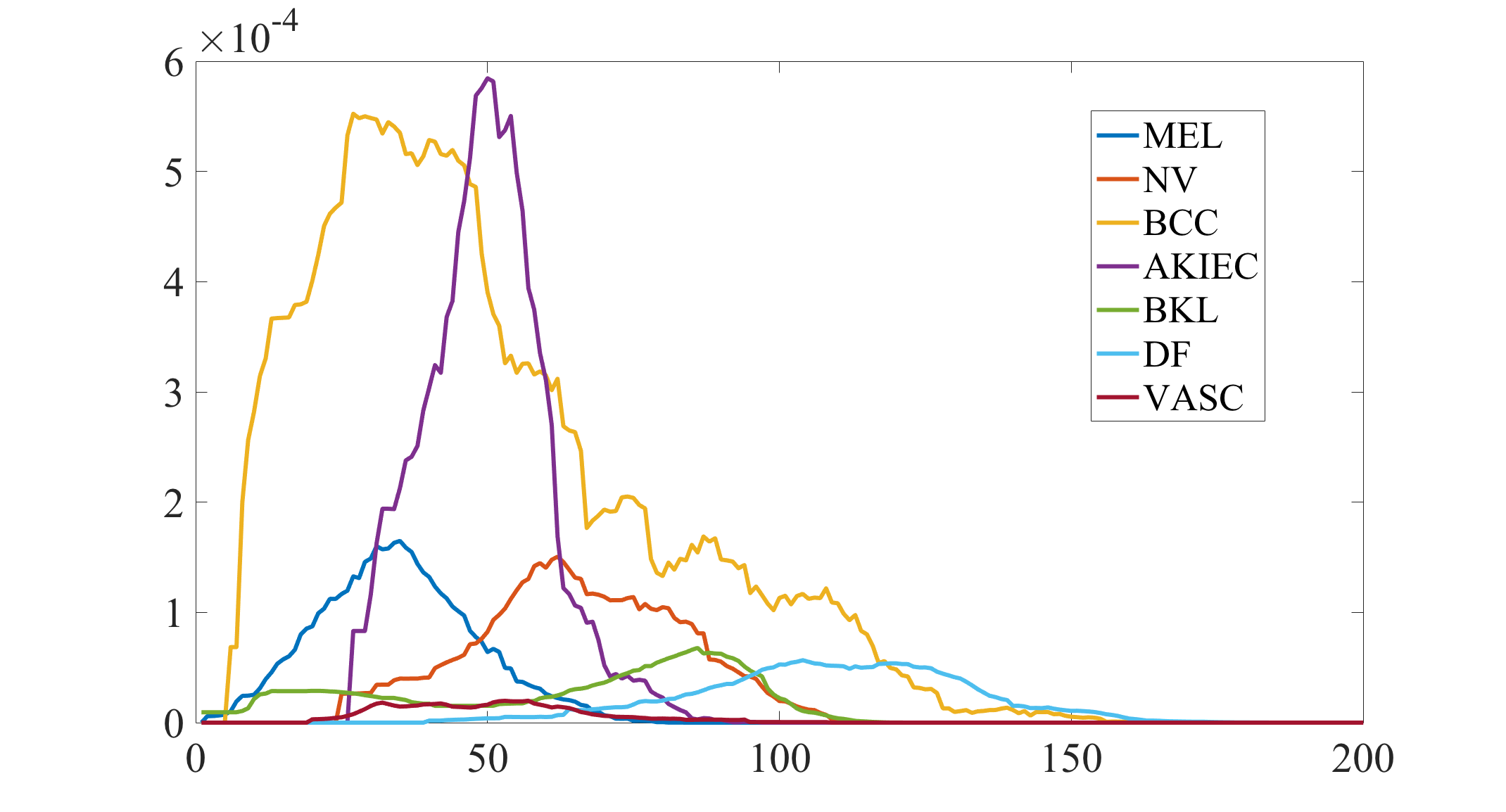}}
\caption{Sample P.Cs from X channel of XYZ color space of images in Fig.~\ref{fig:sample images}.}
\label{fig:sample PC}
\end{figure}

{
On the other hand, PCs offer a method to create a vectorized summary of a persistence diagram. By mapping a persistence diagram to a vector space, we open the door to the application of machine learning algorithms. The motivation for this class of curves lies in the Fundamental Lemma of Persistent Homology which states that given a $k$-dimensional persistence diagram $P$ corresponding to a filtration $X_1\subset \ldots X_n$, the space in the filtration, $X_t$ corresponding to threshold $t$ has exactly $\beta_k(X_t) =|\{(b,d)\in P\mid b\le t, d>t\}|$.  Recall the formal definition of PCs from \cite{ChungLawson}, which is a generalization of this idea.
Let $\mathcal{D}$ represent the set of all persistence diagrams. let $\mathcal{F}$ represent the set of all functions $\psi:\mathcal{D}\times \mathbb{R}^3\to\mathbb{R}$ so that $\psi(D;x,x,t)=0$ for all $x\in\mathbb{R}$. Let $\mathcal{T}$ represent the set of \textbf{statistics} or operators that map multi-sets to the reals, and finally let $\mathcal{R}$ represent the set of functions on $\mathbb{R}$. We define a map $P:\mathcal{D}\times\mathcal{F}\times\mathcal{T}\to\mathcal{R}$ where  \[P(D,\psi, T)(t) = T(\{\psi(D;b,d, t)\mid~ b\le t,~ d > t \}).\] The function $P(D,\psi, T)$ is called a \textbf{persistence curve} on $D$ with respect to $\psi$ and $T$. In \cite{ChungLawson}, it is shown that persistence landscapes \cite{bubenik2015statistical} are a special case of PCs. 

In the present application, all filtrations have exactly 255 space. Thus, for each diagram, a persistence curve is a vector in $\mathbb{R}^{255}$. The two particular functions that were of greatest use were the functions $\psi(b,d,t) = 1$ giving rise to the Betti curve $\beta(t)$ and  $e(b,d,t) = -\frac{d-b}{L}\log \frac{d-b}{L}$ giving rise to a variant of the entropy summary (curve)  $E(t)$. The entropy summary and its stability are discussed in \cite{PersistentEntropy,persistentEntropyStability}. In \cite{ChungLawson}, a general stability result for an entire class of PC is given. We calculate the curves for the 0 and 1 dimensional persistence diagrams for each channel in our color space. Finally, we fed these features into machine learning models. The persistence curves we used in our final model are}
\begin{enumerate}
\item $\beta_0(t)$ and $\beta_1(t)$.
\item $E_0(t)$ and $E_1(t)$.
\end{enumerate}
{In our implementation, these features are described in the following
\begin{itemize}
    \item PS-RGB (dimension = $19\times 3 \times 2 = 114$);
    \item PS-XYZ (dimension = $19\times 3 \times 2 = 114$);
    \item PC-RGB (dimension = $255\times 3 \times 2 = 1530$);
    \item PC-XYZ (dimension = $255\times 2 \times 2 = 1020$);
\end{itemize}
as an input for topological ResNet-101. Note that for both PS-RGB, and PS-XYZ, each channel produces two persistence diagrams (0th and 1st level), and each persistence diagrams summarizes to PS as a $19$-dimensional vector.  Therefore, both PS-RGB and PS-XYZ are of dimension 114.  PC-RGB contains 6 PCs in total, which are $\beta_0(t)$ and $\beta_1(t)$ for each channel.  On the other hand, empirically, we found that in the XYZ color space, X component seems to perform well.  Hence, PC-XYZ contains four PCs in the X channel: $\beta_0(t)$, $\beta_1(t)$, $E_0(t)$, and $E_1(t)$ } Fig.~\ref{fig:sample PC} illustrates samples of persistence curves.  

We summarize our approach as follows. First, we apply the segmentation algorithm discussed in Section \ref{sec:seg} to obtain the image mask.  Second, we apply the mask to the original image.  Third, we transform the RGB color space into RGB, HSV, or XYZ color space, and extract each channel.  Fourth, we use persistent homology software, specifically, {\tt{Perseus}} \cite{Perseus} and {\tt{CubicalRipser}} \cite{CubicalRipser}, to compute persistence diagrams for each channel.  Finally, from each persistence diagram, we calculate persistence curves, and persistence statistics as features. 
\subsection{Segmentation}
\label{sec:seg}
We now describe the segmentation algorithm.  The goal is to take an image $I$ of a patch of skin that has a (contiguous) skin lesion and isolate the connected component of pixels corresponding to the diseased skin.  The output is to be a mask $M$, a binary image with $M(x,y) = 255$ (or white) if pixel $(x,y)$ is part of the lesion and $M(x,y) = 0$ (or black) if pixel $(x,y)$ is part of the healthy skin in the image.

Naturally this is a subjective task.  As a check, a mask $M$ that is obtained from an image $I$ is compared to the mask $M'$ that is obtained by a dermatologist.  The evaluation metric is the Intersection over Union (IOU) score, namely $J(M,M') = |M \cap M'|/|M \cup M'|$.  

The segmentation algorithm proceeds in the following steps.

\noindent {\bf Step 1:} We transform each RGB image $I$ into a gray image $I^*$ that is the average the color intensities, i.e. 
\begin{equation}
I^*(x,y) = \frac{1}{3} \left(R(x,y) + G(x,y) + B(x,y)\right).
\end{equation}

\noindent {\bf Step 2:} Because the region of skin in the lesion is usually darker than the healthy region, we first compute the average value $a$ of $I^*$. If the pixel value in $I^*$ is less than $a$ we take this to mean that it is more likely to be a part of the lesion. The second step is to observe the life interval of each pixel. Like persistent homology, for each step $t$ with $0 \leq t \leq T$, we define $I_t^*$ to be the binary image
\begin{equation}
I_t^*(i,j) = \begin{cases} 
0 				&\mbox{if \ } I^*(i,j) > a \cdot (1 - \frac{t}{T}), \\ 
255 			&\mbox{if \ } I^*(i,j) \leq a \cdot (1 - \frac{t}{T}, 
\end{cases}
\end{equation}
Because pixels in $I_t^*$ usually become $0$ when $t \geq 30$, we choose $T=50$ in our implementation.

\begin{figure}[htp]
\setlength{\abovecaptionskip}{1pt} 
\setlength{\belowcaptionskip}{1pt} 
\centering 
        \includegraphics[width=\linewidth]{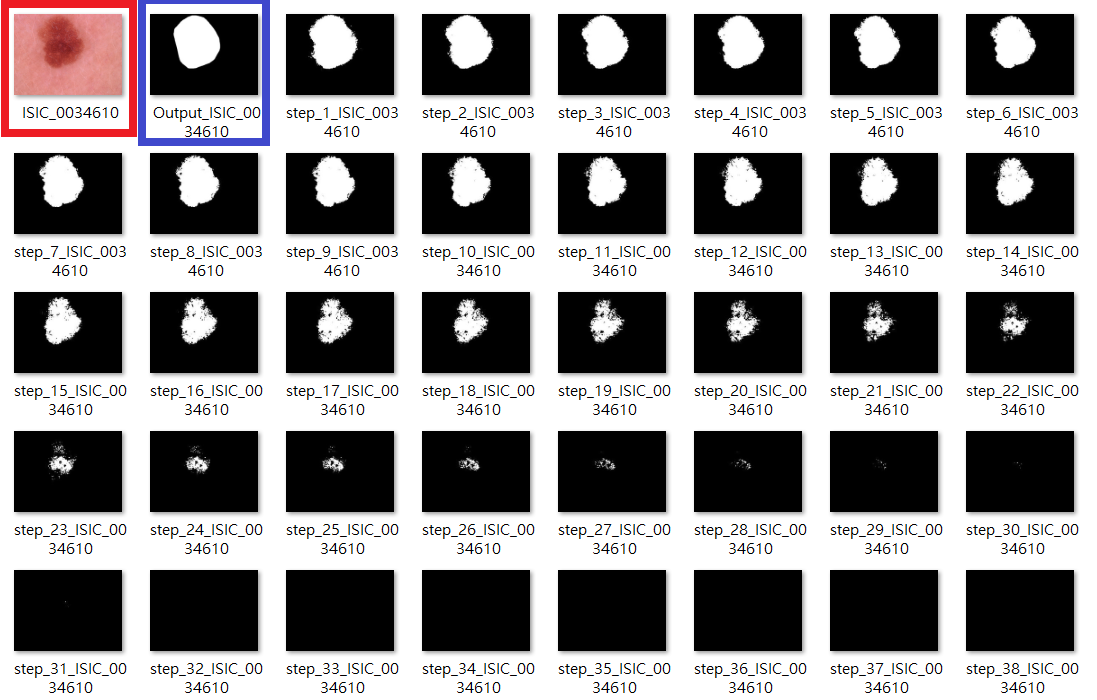}
        \caption{An example of our main idea in proposed algorithm. The images bounded by red and blue bounding boxes are original skin image and output segmentation respectively. The other images are $S_1 \sim S_{38}$ in \eqref{Equation : Filtration}. In this class, our segmentation algorithm selects $t = 2$ by equation \eqref{Equation : Criterion for choosing threshold}, and the output segmentation would be the convex hull of the main connected component in $S_2$, which is derived by \eqref{Equation : Life score formula}.}
        \label{Figure : Steps}
\end{figure}

We define $S_t$ to be the set of all white pixels in $I_t^*$ in step $t$ and get the following filtration:  
\begin{equation}
\label{Equation : Filtration}
S_0 \supseteq S_1 \supseteq S_2 \supseteq S_3 \supseteq \cdots \supseteq S_T,
\end{equation}
This filtration is illustrated in Fig.~\ref{Figure : Steps} for one training image. As in persistent homology, we measure the life interval of each white pixel in $S_1$, white pixels in $S_1$ with long life intervals have a more robust property in the whole image, and are more likely to be a part of the lesion. If $(x,y) \in S_t$ for $1 \leq t \leq t_0$ and $(x,y) \not \in S_t$ for $t > t_0$ we say the life-span of $(x,y)$ is $L(x,y) = t_0$.  If $(x,y) \not \in S_1$ we set $L(x,y)=0$.

\noindent {\bf Step 3:} For each image, we calculate a threshold $T'$ with $1 < T' < T$.  The output segmentation is determined by connected components in $S_{T'}$. 

 
In general, as $t$ gets larger, the number of connected components will decrease since noisy parts would disappear first. However, when $t$ gets very large, the main components would be separated by small components. Therefore, we choose

\begin{equation}
\label{Equation : Criterion for choosing threshold}
T' = 1 + \left \lfloor \frac{T''}{4} \right \rfloor,
\end{equation}
where $1 \leq T'' \leq T$ is the first time step such that $S_{T'' + 1}$ has more connected components than $S_{T'' + 1}$, and $4$ is a choice to ensure the main disease component is more complete.

\noindent {\bf Step 4:} A binary image $S_{T'}$ with small $T'$ usually contains many tiny (or noisy) white components, so we remove them and define life scores $\rm LS$ of connected components of $S_{T'}$. More precisely, if $C$ is a (white) connected component of $S_{T'}$, then its life score is defined by modified average life-span
\begin{equation}
\label{Equation : Life score formula}
{\rm LS}(C) = \frac{(1 + d(C, b))^3 \cdot \sum_{(x,y) \in C} L(x,y)}{(1 + d(C, o))^3},
\end{equation}
where $(x,y)$ is a pixel in $C$ and $d(C,o)$, $d(C,b)$ are the minimal distance between $c$ and midpoint $o$ and boundary of the image respectively. Both of $1+ d(C,b)$ and $1+ d(C,o)$ factors occur in \eqref{Equation : Life score formula} as punishments on components not near the center of the image. (Usually the lesion appears close to the center of the image.) A connected component $C$ with higher score means that this region might be more significant in the whole image. Thus we choose the convex hull of those connected components $C_1, C_2, \dots, C_k$ with life scores larger than the average life score among all connected components.

The proposed method was considered by topological properties of raw images purely, hence any training process were not required. The average IOU score among $2595$ images we obtained was $0.6651$.



\section{Topological ResNet-101}
\label{Sec: Top ResNet}
{
Convolution neural networks (CNN) become main tools in the deep learning.  Many researchers have been developing several CNN models, such as (just to name a few) AlexNet~\cite{AlexNet}, VGG~\cite{VGG} and ResNet~\cite{resNet}.  Applications of those deep learning models to the computer vision have proven successful, such as \cite{Fully-Net,U-Net,resNet}.  In this work, we will focus on the residual neural network (ResNet);  in particular, we will use the ResNet-101~\cite{resNet}, which provides an end-to-end architecture for the image classification.  ResNet optimizes the residues between input and desired convolution features.  The desired features can be extracted easier and more efficient than other CNN models. Therefore, this optimization of residues can be applied to reducing the number of parameters in a heavy network. Because of the benefit of reduction of parameters, the number of layers can be lifted efficiently. In this work, we modified ResNet with 101 layers by equipping the topological information.



\begin{figure*}
\includegraphics[width=\linewidth]{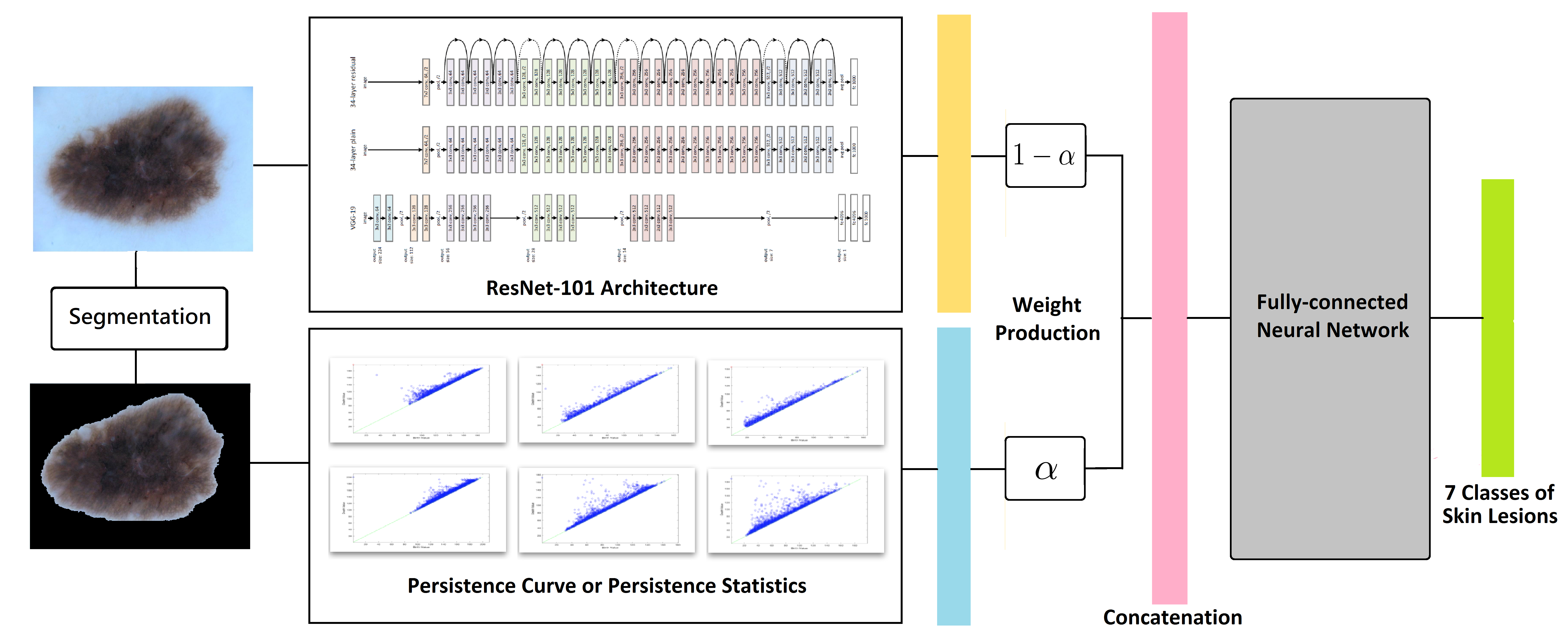}
\caption{The architecture of the TopoResNet-101, where the blue bar is PS, PC or their concatenations and $\alpha \in [0,1]$ is the proposed topological rate. The yellow bar is the output vector of ResNet-101 before fully-connected layer.}
\label{fig: Architechture}
\end{figure*}
The architecture of the TopoResNet-101 can be shown in Fig. \ref{fig: Architechture}. We introduce a new parameter $\alpha\in [0,1]$, called the {\it topological rate}, as a weight of different features.  We multiply each component in topological features (ResNet-101 output features) by $\alpha$ ($1-\alpha$, respectively).  To be more specifically, the input vector before the last fully-connected in the TopoResNet-101 i.e., the pink bar in Fig. \ref{fig: Architechture} can be represented by
\begin{equation}
\label{Equation : alpha concatenation}
    \mathbf{v} = (1-\alpha) \cdot \mathbf{v}_{\rm ResNet-101} \oplus \alpha \cdot \mathbf{v}_{\rm Topology},
\end{equation}
where $\mathbf{v}_{\rm ResNet-101}$ is the ResNet-101 output features (yellow bar in Fig.~\ref{fig: Architechture}, and $\mathbf{v}_{\rm Topology}$ is the topological features (blue bar in Fig.~\ref{fig: Architechture}), and $\oplus$ is the concatenation operator.  Since $\alpha$ is a parameter in the TopoResNet-101, it will be changed in the learning process.
In practice, $\alpha$ was initially set to be $\sigma(0.5) \approx 0.6$, where $\sigma : \mathbb{R} \rightarrow [0,1]$ is the sigmoid function
\begin{equation*}
    \sigma(t) = \frac{1}{1 + e^{-t}}.
\end{equation*} 
This sgimoid function will ensure that the $\alpha$ is always in between 0 and 1.  In TopoResNet-101, we will always apply the sigmoid function to $\alpha$.
By regarding ResNet-101 as a regression model of data points, $\alpha$ records the importance of topological features for classifying skin lesions. We also allow $\alpha$ as a weight in the network, so it can be optimized in training epochs. Fig. \ref{fig: Topological rates of Model 4,5,6,7} and \ref{fig: Topological rates} show the role of $\alpha$ in the training process for each model.   

}

\section{Experiments}
\label{Sec:Exp}
{
In this section, we describe the our main experiments and results.  There are two models in this work---SVM and TopoResNet-101.  Since the experimental settings of SVM model and TopoResNet-101 are slightly different, we explain the differences between those settings in Section \ref{subsec : data description}.  In Section \ref{sec:Performance of SVM} and \ref{sec:Performance of TopResNet}, we present the performances of SVM and  TopoResNet-101, respectively.  The affects of the topological rate, $\alpha$, are discussed in \ref{sec:Topological rates}.
}

\subsection{Data Description and Experiment Setting}
\label{subsec : data description}
\begin{table}\renewcommand{\arraystretch}{1.5}
\centering
\caption{Numbers of skin lesion images in each class.}
\begin{tabular}{|c|c|c|}
\hline
No. & Class  & Number \\ \hline
0 & Melanoma (MEL) & 1113  \\ \hline
1 & Melanocytic nevus (NV) & 6705  \\ \hline
2 & Basal cell carcinoma (BCC) &  514  \\ \hline
3 & Actinic keratosis (AKIEC) &  327  \\ \hline
4 & Benign keratosis (BKL) &  1099  \\ \hline
5 & Dermatofibroma (DF) &  115  \\ \hline
6 & Vascular lesion (VASC) &  142  \\ \hline
\end{tabular}
\label{table: data information}
\end{table}

{
As shown in Table~\ref{table: data information}, the distribution of images among classes is extremely unbalanced.  In the SVM model, we use multi-class SVM with a ``one-against-one'' strategy and all 10015 skin lesion images as training input of SVM.  An additional set of images as a validation dataset was provided by ISIC 2018 website as part of the challenge and the class information of validation images were not available to the end users.  Because unbalanced training set may lead to over-fitting on classes which have large cardinality, we train the model by using a subset of the full training set (5000 images chosen randomly from the training set by MATLAB's random seed 1).

On the other hand, because the ISIC 2018 validation system had been closed, we collected $50$ images from each class as testing dataset (totally $350$ images) in original 10015 skin lesion images.  For the training process, we separated $70\%$ of the rest images as training data and the others $30\%$ as the validation data. Because CNN models usually need massive training set, the random sampling of training set is not applied for CNN models; in other words, all 10015 skin lesion images were inputted in training process.



} 

\subsection{Performance of SVM}
\label{sec:Performance of SVM}
The best 2 scores we had on the validation set was $65.6\%$, and $67.2\%$ as shown in Table \ref{tab:main scores}.
\begin{table}\renewcommand{\arraystretch}{1.5}
\caption{Balanced accuracy on validation set by SVM.}
\label{tab:main scores}
\begin{tabular}{|c|c|c|}
\hline
& Features & $\stackrel{\mathrm{Validation}}{\mathrm{Score}}$ \\ \hline
SVM 1& PC-XYZ, PS-XYZ & $65.6\%$ \\ \hline
SMV 2& PC-XYZ, PS-XYZ, and PS-RGB & $67.2\%$ \\ \hline
\end{tabular}
\end{table}
The experiment result shows that global information (PS and PC) may be useful features for classifying skin lesions. However, because local features were not considered in the model. Therefore, this motivates us to combine PS and PC into CNN models.


\subsection{Performance of TopoResNet-101}
\label{sec:Performance of TopResNet}
\begin{figure*}
\subfloat[Class 0 (MEL)]{\includegraphics[scale=0.40]{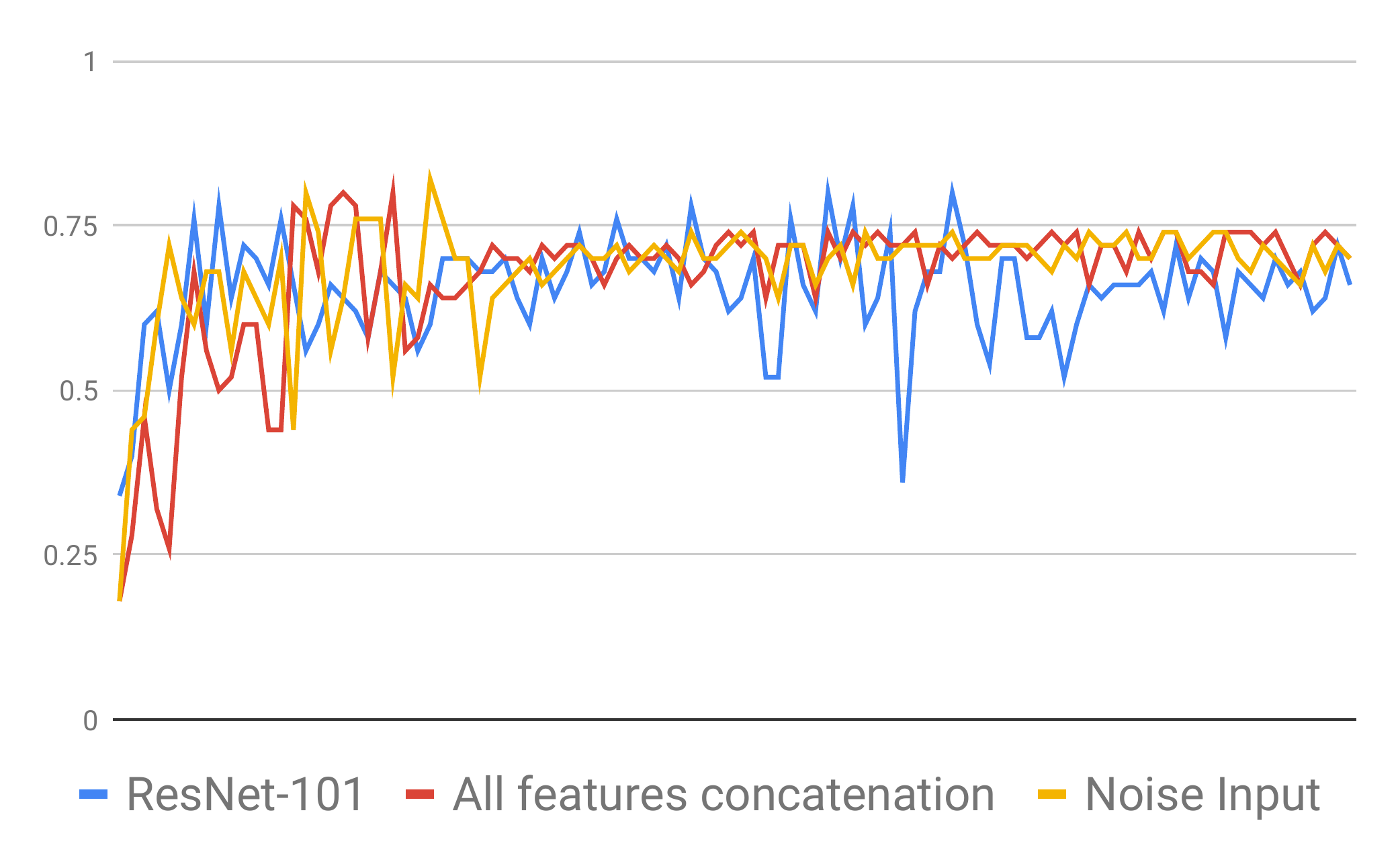}}\hfill
\subfloat[Class 1 (NV)]{\includegraphics[scale=0.40]{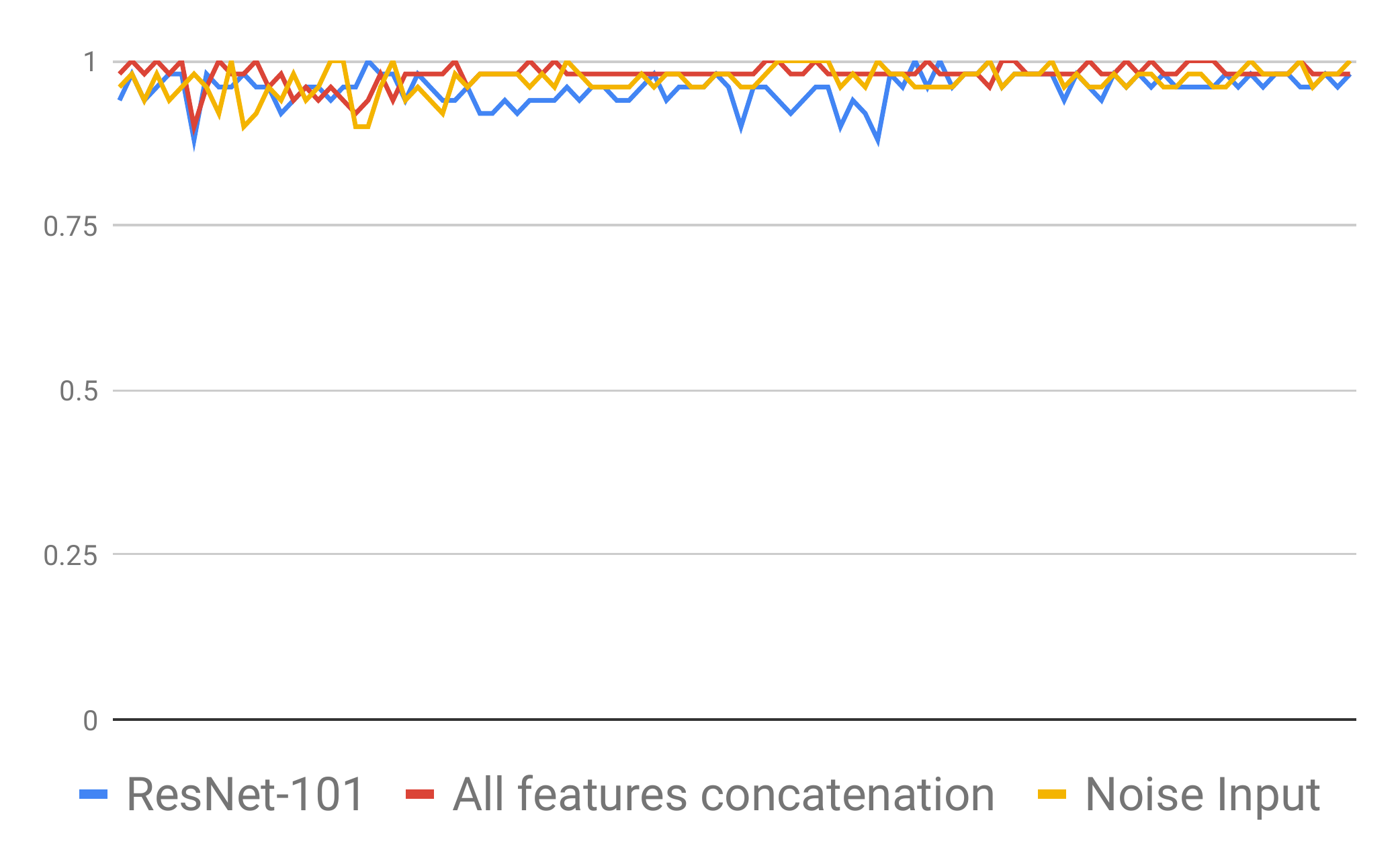}}\hfill
\subfloat[Class 2 (BCC)]{\includegraphics[scale=0.40]{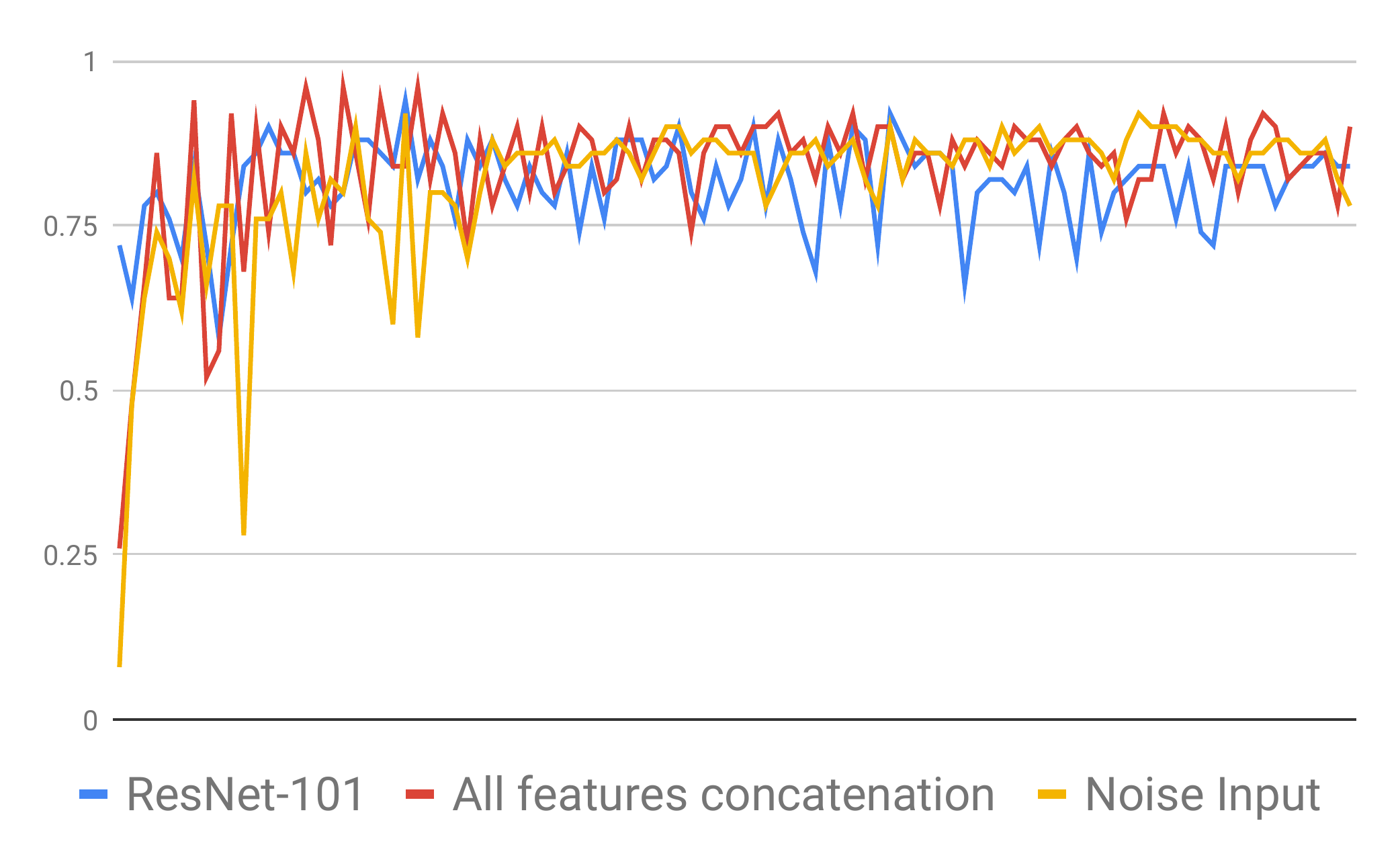}}\hfill
\subfloat[Class 3 (AKIEC)]{\includegraphics[scale=0.40]{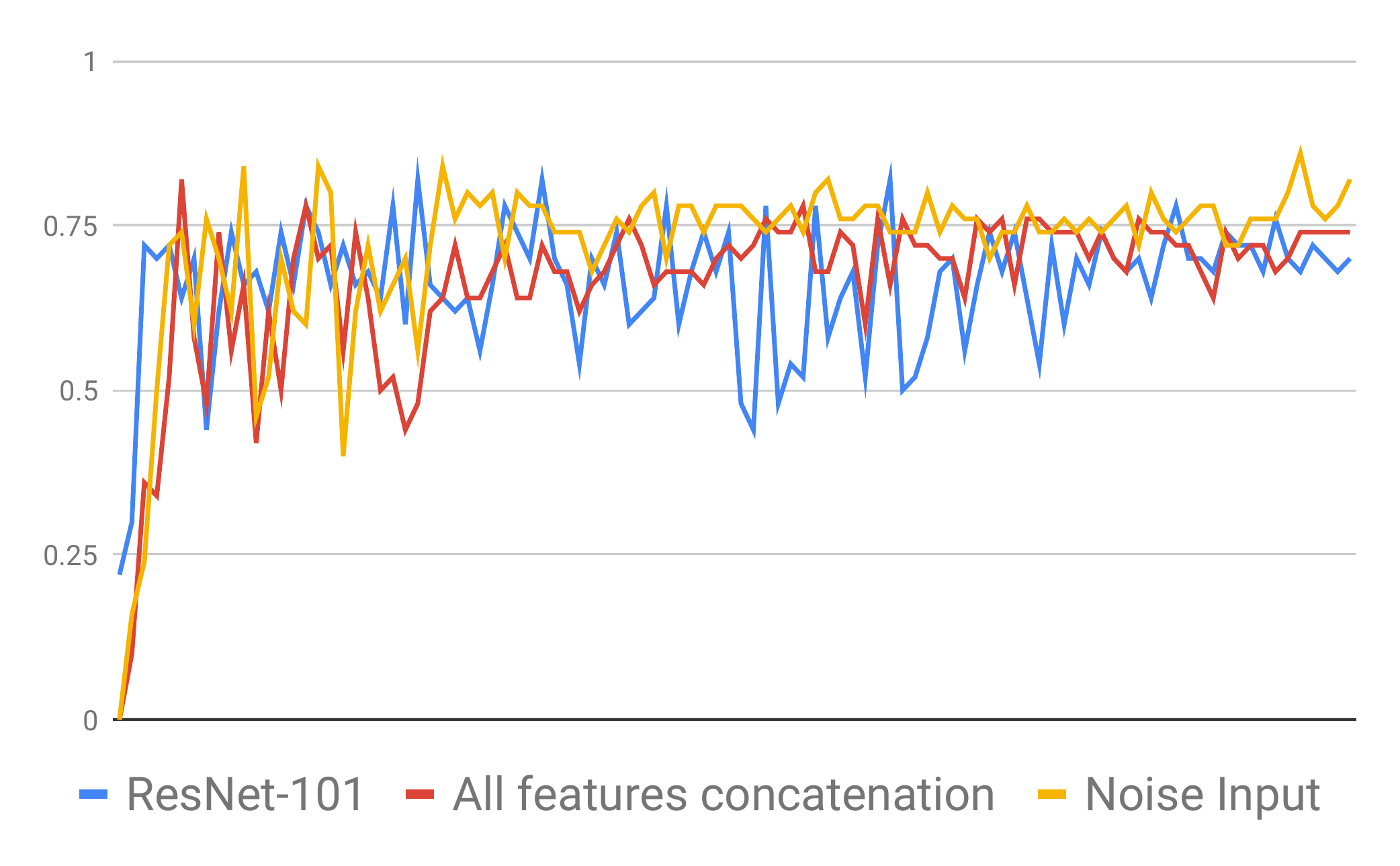}}\hfill
\subfloat[Class 4 (BKL)]{\includegraphics[scale=0.40]{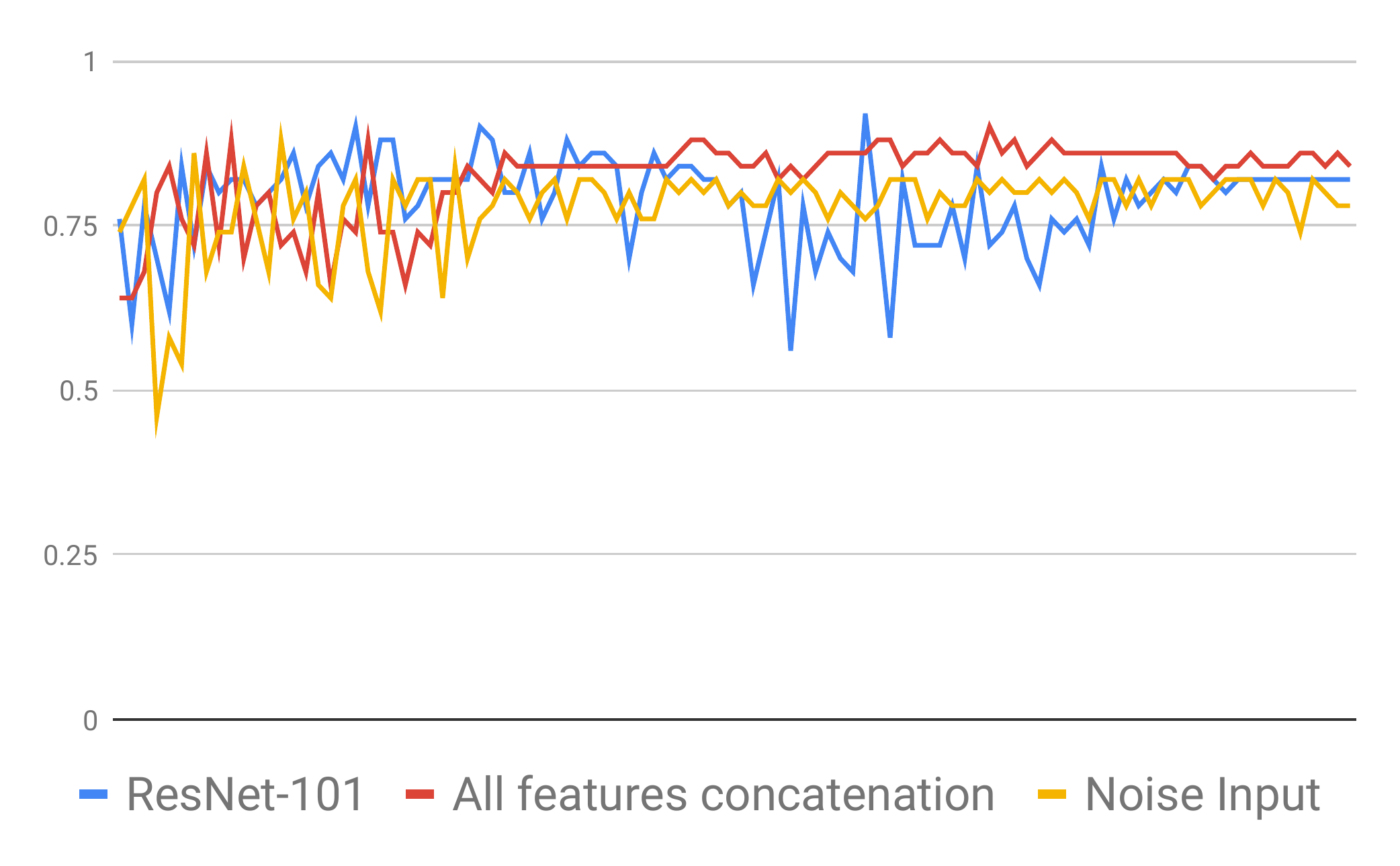}}\hfill
\subfloat[Class 5 (DF)]{\includegraphics[scale=0.40]{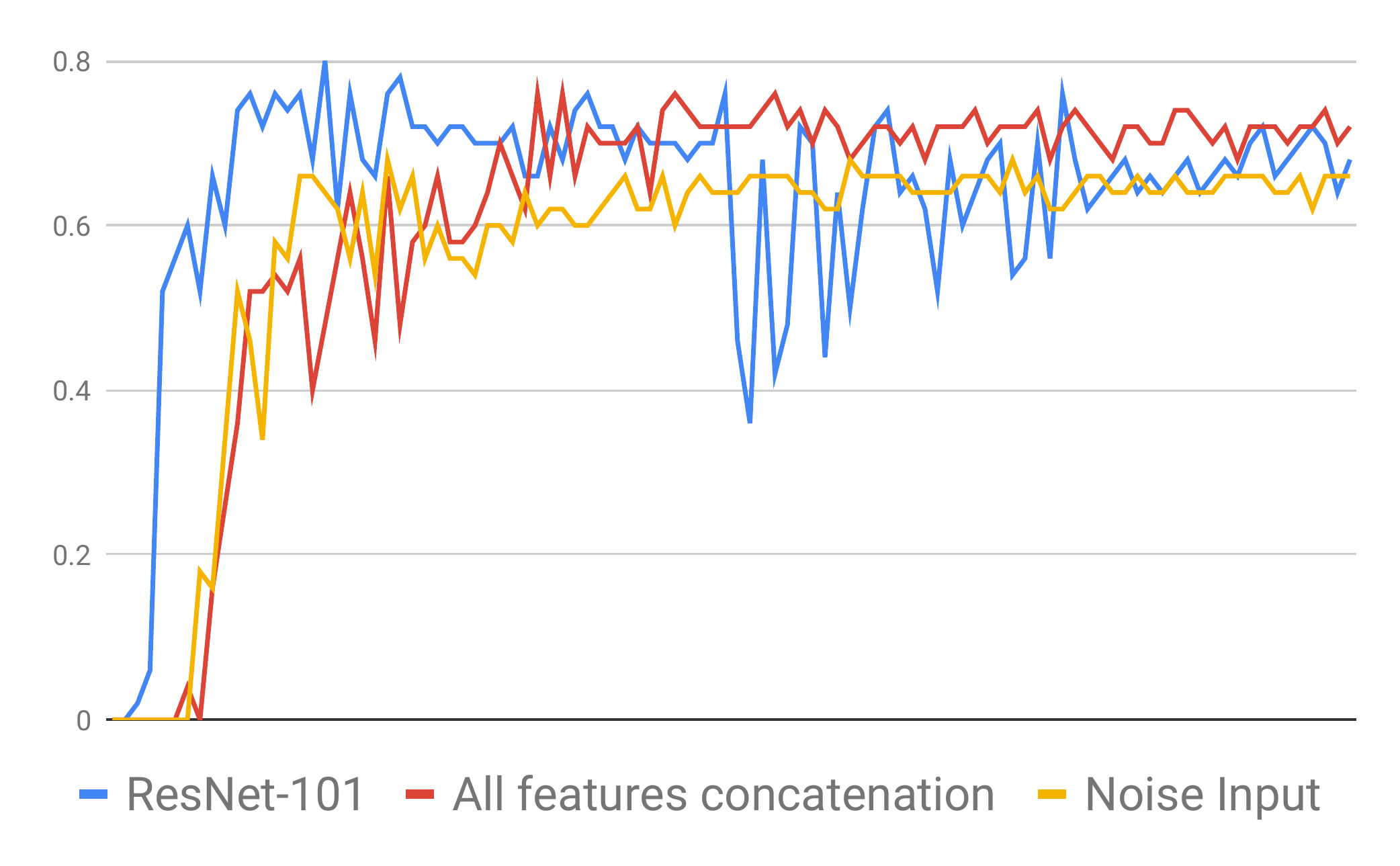}}\hfill
\subfloat[Class 6 (VASC)]{\includegraphics[scale=0.40]{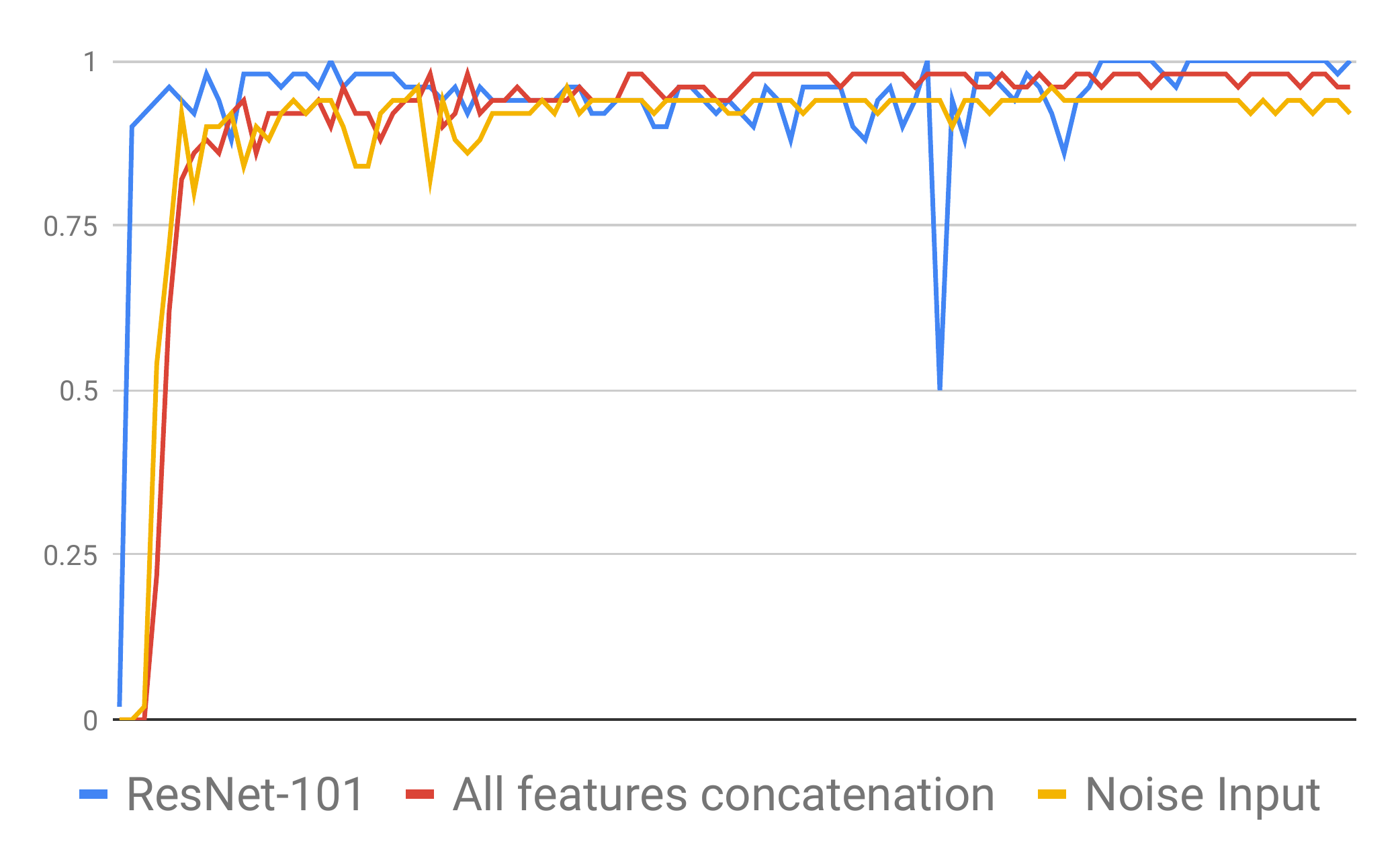}}\hfill
\subfloat[Balanced Accuracy]{\includegraphics[scale=0.40]{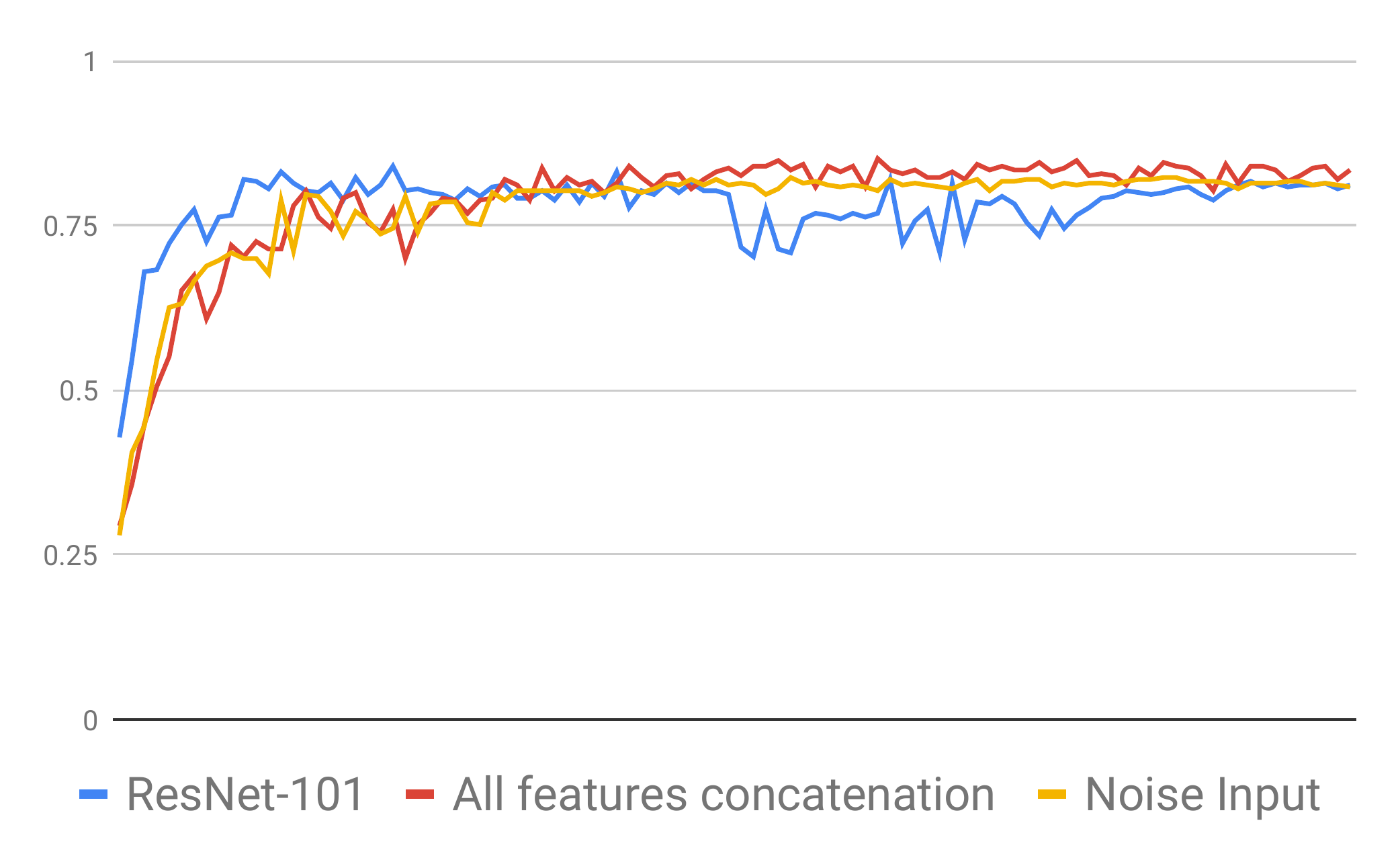}}\hfill
\caption{Mean and classes accuracy curves of Model 1, 8 and 9 on training processes.}
\label{fig:class Acc}
\end{figure*}

{For each topological features (totally $4$ sets of features), we trained corresponding TopoResNet-101. For each model in the experiments, we observe convergence of mean class accuracy curve in training epochs on validation dataset and picked the weights with maximal accuracy when the curve converged before over-fitting on validation set. In our experiments, we performed 9 models for observing the performances of RestNet-101 equipped with different topological features. These models are listed below:
\begin{enumerate}[(Model 1):]
    \item N/A (original ResNet-101);
    \item PS-RGB (dimension = 114);
    \item PS-XYZ (dimension = 114);
    \item PC-RGB (dimension = 1530);
    \item PC-XYZ (dimension = 1022);
    \item Reduced PC-RGB (dimension = 512);
    \item Reduced PC-XYZ (dimension = 512);
    \item Reduced all topological features (dimension = 512);
    \item Noise data (dimension = 512).
\end{enumerate}

Table \ref{tab: Performances of ResNets} presents the performance of the Model $1 \sim 5$ on balanced test dataset. Because features with large dimensions (PC-RGB and PC-XYZ) may result in unstable in the training process, we design a sub-net architecture for reducing the dimension of topological features before concatenation (i.e., pink one in Fig.\ref{fig: Architechture}). Table \ref{tab: Performances of ResNets} presents the performance of the Model $6 \sim 9$, where the blue layer in Fig.\ref{fig: Architechture} was replaced by Fig. \ref{fig: reduction_net}.

\begin{figure}
\includegraphics[width=\linewidth]{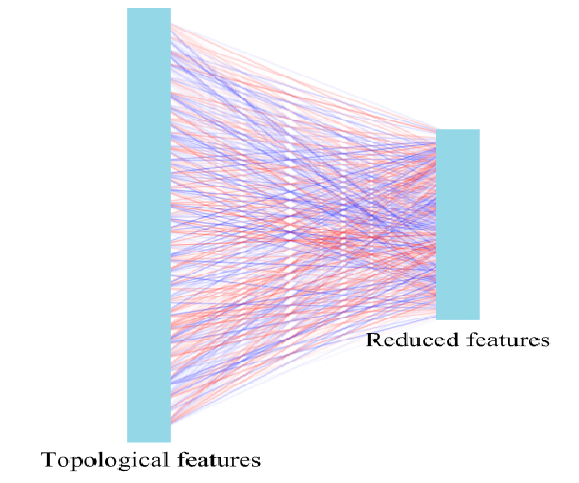}
\caption{ The sub-net for reducing topological features in Model $6 \sim 9$, where the reduced dimension is set to be 512.}
\label{fig: reduction_net}
\end{figure}

The reduced dimension in Fig. \ref{fig: reduction_net} is set to be 512 via a fully-connected sub-net merged into  TopoResNet-101, and the weights of the sub-net were optimized synchronously by back-propagation technique in training processes.

Finally, because adding noise into neural networks may improve the performance~\cite{noise-Assistance}, we also consider 512 dimensional noise vectors as an additional input of ResNet-101 (Model 9). In practice, all input features were normalized into a vector where each component of the vector is a value between $0$ and $1$ by Max-Min normalization.

\begin{table}\renewcommand{\arraystretch}{1.5}
\centering
\caption{Balanced accuracy of ResNet-101 and models of ResNet-101 with different topological features on training set and related topological rate $\alpha$.}
\label{tab: Performances of ResNets}
\begin{tabular}{|c|c|c|c|}
\hline
No. & Model  & Acc. & $\alpha$ \\ \hline
1 & ResNet-101 & 0.806 & N/A \\ \hline
2 & ResNet-101 + PS-RGB & \textbf{0.84} & 0.152 \\ \hline
3 & ResNet-101 + PS-XYZ &  0.831 & 0.199 \\ \hline
4 & ResNet-101 + PC-RGB &  0.76 & 0.366 \\ \hline
5 & ResNet-101 + PC-XYZ &  0.783 & 0.318 \\ \hline
\end{tabular}
\end{table}

\begin{table}\renewcommand{\arraystretch}{1.5}
\centering
\caption{Balanced accuracy of ResNet-101 models combinied with reduced PC-RGB, reduced PC-XYZ, concatenation of all features and random noise input.}
\label{tab: Performances of ResNets (with feature Reduction)}
\begin{tabular}{|c|c|c|c|}
\hline
No. & Model  & Acc. & $\alpha$ \\ \hline
6 & ResNet-101 + PC-RGB (reduced) & 0.837 & 0.22 \\ \hline
7 & ResNet-101 + PC-XYZ (reduced) & 0.837 & 0.21 \\ \hline
8 & ResNet-101 + all features (reduced) & \textbf{0.851} & 0.23 \\ \hline
9 & ResNet-101 + random noise (reduced) & 0.814 & 0.191 \\ \hline
\end{tabular}
\end{table}

By observing Table \ref{tab: Performances of ResNets} and Table \ref{tab: Performances of ResNets (with feature Reduction)}, the results show that Model 8 has the best performance among these 9 models. To measure the effect of topological features in neural networks, we plotted mean accuracy and class accuracy of Models 1, 8 and 9 in Fig. \ref{fig:class Acc} in training processes on test dataset. By observing curves in Fig. \ref{fig:class Acc}, Model 8 has most stable performance on testing dataset. As we mentioned in Section \ref{subsec : data description}, noise embedding in neural networks would also makes the neural network performs more robust. However, except AKIEC lesions (Class 3), topological features performs well than noise inputs significantly. 
}
\subsection{$\alpha$ Rates in TopoResNet-101}
\label{sec:Topological rates}

\begin{figure}
\includegraphics[width=\linewidth]{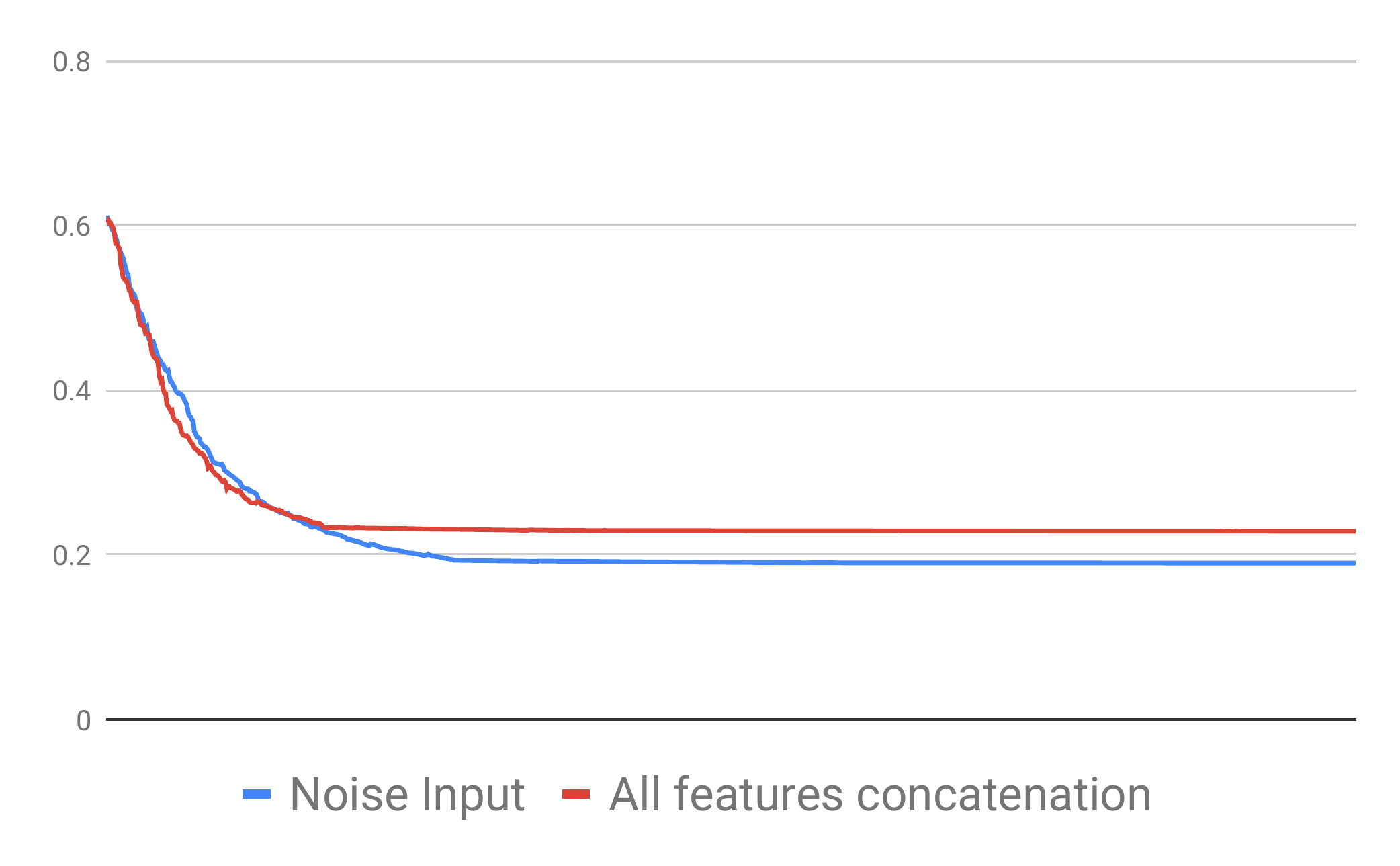}
\caption{Topological rates $\alpha$ of Model 8 and Model 9. The horizontal line is the epoch number in training processes.}
\label{fig: Topological rates}
\end{figure}

{First, we note that $\alpha$ converges in all models as shown in Fig.~\ref{fig: Topological rates} and \ref{fig: Topological rates of Model 4,5,6,7}.  Observe from Table \ref{tab: Performances of ResNets} that the $\alpha$ rate seems to be influenced by the dimension of input features as Model 4 and 5 have higher feature dimension that Model 2 and 3 do.  This is one of reasons we consider Model 6, 7, and 8.  In Fig.~\ref{fig: Topological rates}, $\alpha$ rates converge in training processes, and the $\alpha$ rate of Model 8 is lager than Model 9, this phenomenon suggests that PS and PC are useful for recognizing skin lesions.

}

{
On the other hand, to investigate the effects of dimension reduction, we also plotted $\alpha$ curves of Model 4, Model 5, Model 6, Model 7 in Fig. \ref{fig: Topological rates of Model 4,5,6,7}. It shows that the reduction network does help optimize $\alpha$ rate more efficiently. However, observe that curves of Model 4 and Model 5 (yellow curve and green curve) in Fig. \ref{fig: Topological rates of Model 4,5,6,7} also show significant peaks and increasing of $\alpha$ rates in earlier epochs. This may suggest that some global information in PC-RGB and PC-XYZ might be lost in the unbalanced training dataset. 
}

\begin{figure}
\includegraphics[width=\linewidth]{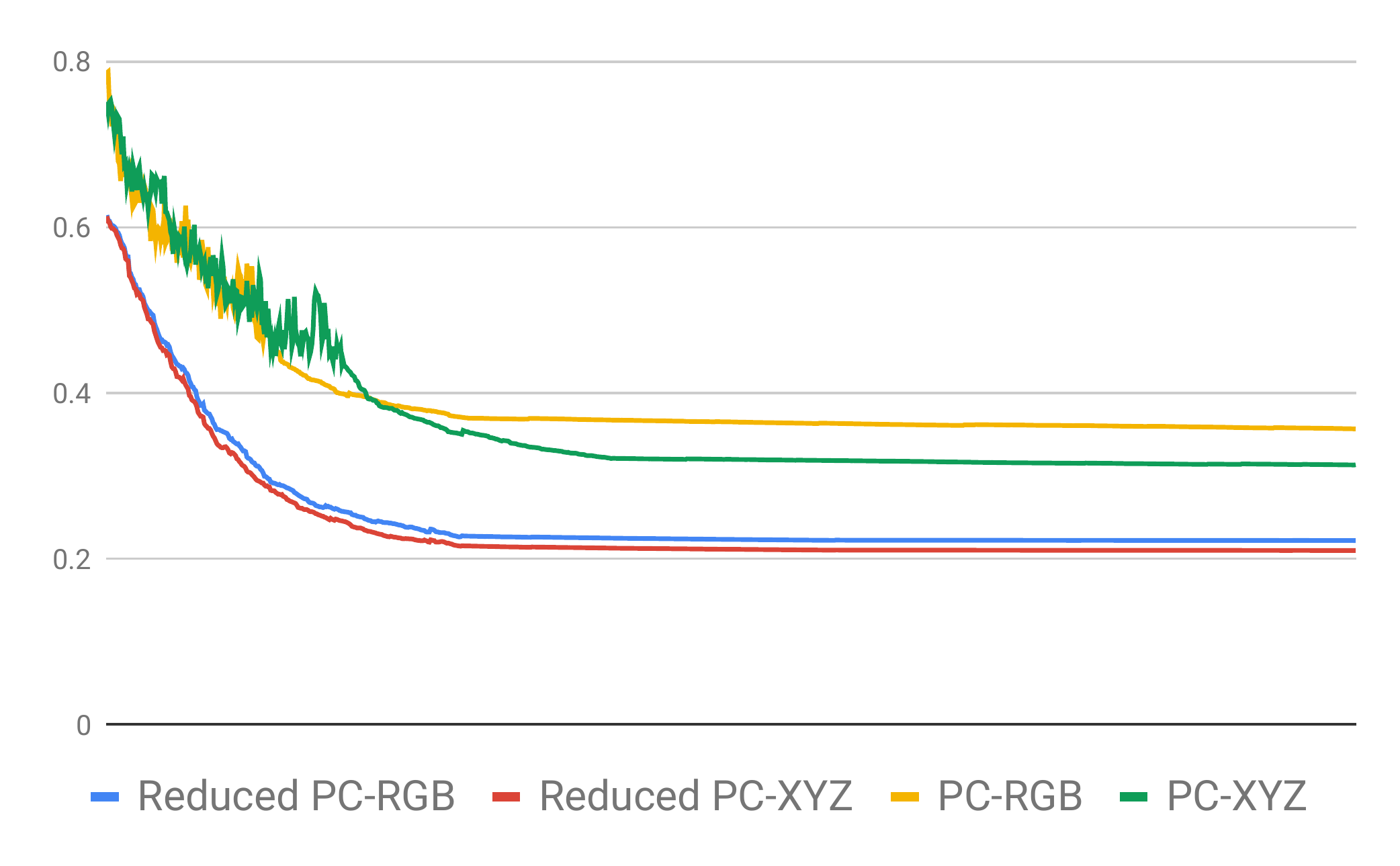}
\caption{Topological rates $\alpha$ of Model 4, 5, 6 and 7. The horizontal line is the epoch number in training processes.}
\label{fig: Topological rates of Model 4,5,6,7}
\end{figure}

\subsection{Discussion}
{In the deep learning, choosing weights of models is a diffcult task, and it is still an open question whether there is a mathematical/theoretical algorithm or criterion to determine the optimal weight in training epochs. Moreover, because a typical method for choosing weights strongly depends on the performance of validation sets, the accuracy between testing and validation datasets may have a gap if the validation set is unbalanced or which has small amount, while this situation occurs frequently in medical images and data. Therefore, the stability of performance is definitely an advantage for choosing models. In our proposed method, neural models equipped with (reduced) topological features have most stable and good accuracy curve on classification task.
}

\section{Conclusion and future work}
\label{sec:conclusion}
The appeal of the PCs and PSs lie in their simplicity. The features themselves do not require user defined parameters, thus one only needs to tune the attached machine learning algorithm. In addition, these features give intuitive shape summaries of the original space. The generalized nature of the persistence curve definition allow for a rich library of usable curves. In this paper, we have chosen to combine the PSs with the Betti and entropy curves and then feed them into {SVM and ResNet-101. The performance shows that PS and PC can be used for lifting the models who considered convolution features only. One future direction is to apply these features into other classification tasks. Also, phenomenon in Fig. \ref{fig: Topological rates} and \ref{fig:class Acc} shows that TopoResNet-101 with high $\alpha$ rate may perform well in specific classes ({\em Class 2}), so it may exist better way to embed $\alpha$ rate to neural models. Finally, because topological features were assumed to be stable on image with noise and it may occurs frequently in real application, it is another important future work to extend the data set with noise and consider the performances between pure ResNet-101 and TopoResNet-101. 
}  


$ $\\
$ $\\

\section*{Acknowledgments}
YMC. would like to thank National Center for Theoretical Science (NCTS), Taiwan for their kindly host.  Much of this work was done during his visit at NCTS in Summer 2018.

\bibliographystyle{IEEEtran}
\bibliography{references}

\end{document}